\documentclass[10pt,twocolumn,letterpaper]{article}

\usepackage[accsupp]{axessibility} % Improves PDF readability for those with disabilities.
\usepackage[article,algorithms]{wacv}      % To produce the REVIEW version for the algorithms track
\usepackage{graphicx}
\usepackage{amsmath}
\usepackage{amssymb}
\usepackage{booktabs}
\usepackage{times}
\usepackage{epsfig}
\usepackage{utfsym}
\usepackage{threeparttable}
\usepackage{multirow} 
\usepackage{makecell}
\usepackage{braket}
\usepackage{pifont}
\usepackage{bbding}
\usepackage{bbm}
\usepackage{array}

\usepackage[pagebackref,breaklinks,colorlinks]{hyperref}

\usepackage[capitalize]{cleveref}
\crefname{section}{Sec.}{Secs.}
\Crefname{section}{Section}{Sections}
\Crefname{table}{Table}{Tables}
\crefname{table}{Tab.}{Tabs.}

\def\wacvPaperID{600} % *** Enter the WACV Paper ID here
\def\confName{WACV}
\def\confYear{2025}

\begin{document}

\title{LiCamPose: Combining Multi-View LiDAR and RGB Cameras for Robust Single-timestamp 3D Human Pose Estimation
}

% \and

\author{Zhiyu Pan \quad  Zhicheng Zhong \quad  Wenxuan Guo \quad  Yifan Chen \quad  Jianjiang Feng\thanks{Jianjiang Feng is the corresponding author. \\ \indent This work was supported in part by the National Natural Science Foundation of China under Grant 62376132 and 62321005.} \quad  Jie Zhou \\
{Department of Automation, BNRist, Tsinghua University, China} \\
{\tt\small \{pzy20, zhongzc18, gwx22, chenyf21\}@mails.tsinghua.edu.cn} \\
{\tt\small \{jfeng, jzhou\}@tsinghua.edu.cn}
}
% \thanks{Corresponding author} # BNRist, 

\maketitle
%%%%%%%%% ABSTRACT
\begin{abstract}
   Several methods have been proposed to estimate 3D human pose from multi-view images, achieving satisfactory performance on public datasets collected under relatively simple conditions. However, there are limited approaches studying extracting 3D human skeletons from multimodal inputs, such as RGB and point cloud data. To address this gap, we introduce LiCamPose, a pipeline that integrates multi-view RGB and sparse point cloud information to estimate robust 3D human poses via single timestamp. We demonstrate the effectiveness of the volumetric architecture in combining these modalities. Furthermore, to circumvent the need for manually labeled 3D human pose annotations, we develop a synthetic dataset generator for pretraining and design an unsupervised domain adaptation strategy to train a 3D human pose estimator without manual annotations. To validate the generalization capability of our method, LiCamPose is evaluated on four datasets, including two public datasets, one synthetic dataset, and one challenging self-collected dataset named BasketBall, covering diverse scenarios. The results demonstrate that LiCamPose exhibits great generalization performance and significant application potential. The code, generator, and datasets are available at https://github.com/Yu-Yy/LiCamPose.
\end{abstract}

\section{Introduction}
Human pose estimation is a fundamental task in computer vision and has been widely applied in various fields, such as human-computer interaction, human activity recognition, sports analytics, augmented reality, etc. Specifically, multi-view image datasets \cite{human36m,panoptic,3dps,3dpw} allow more precise 3D human pose estimation compared to single-view ones \cite{mpii,pose2seg,COCO,lin2014microsoft}, due to the ability of multiple views to capture 3D information from epipolar geometry. As technology advances, researchers \cite{2019learnable,2020voxelpose,2021multi,2021direct,2021tessetrack} have achieved promising results on current public multi-view images datasets. However, practical scenarios are more challenging than existing public datasets, with diverse human motions, severe occlusions, and large scene.  

\begin{figure}
   \centering
   \includegraphics[width=\linewidth]{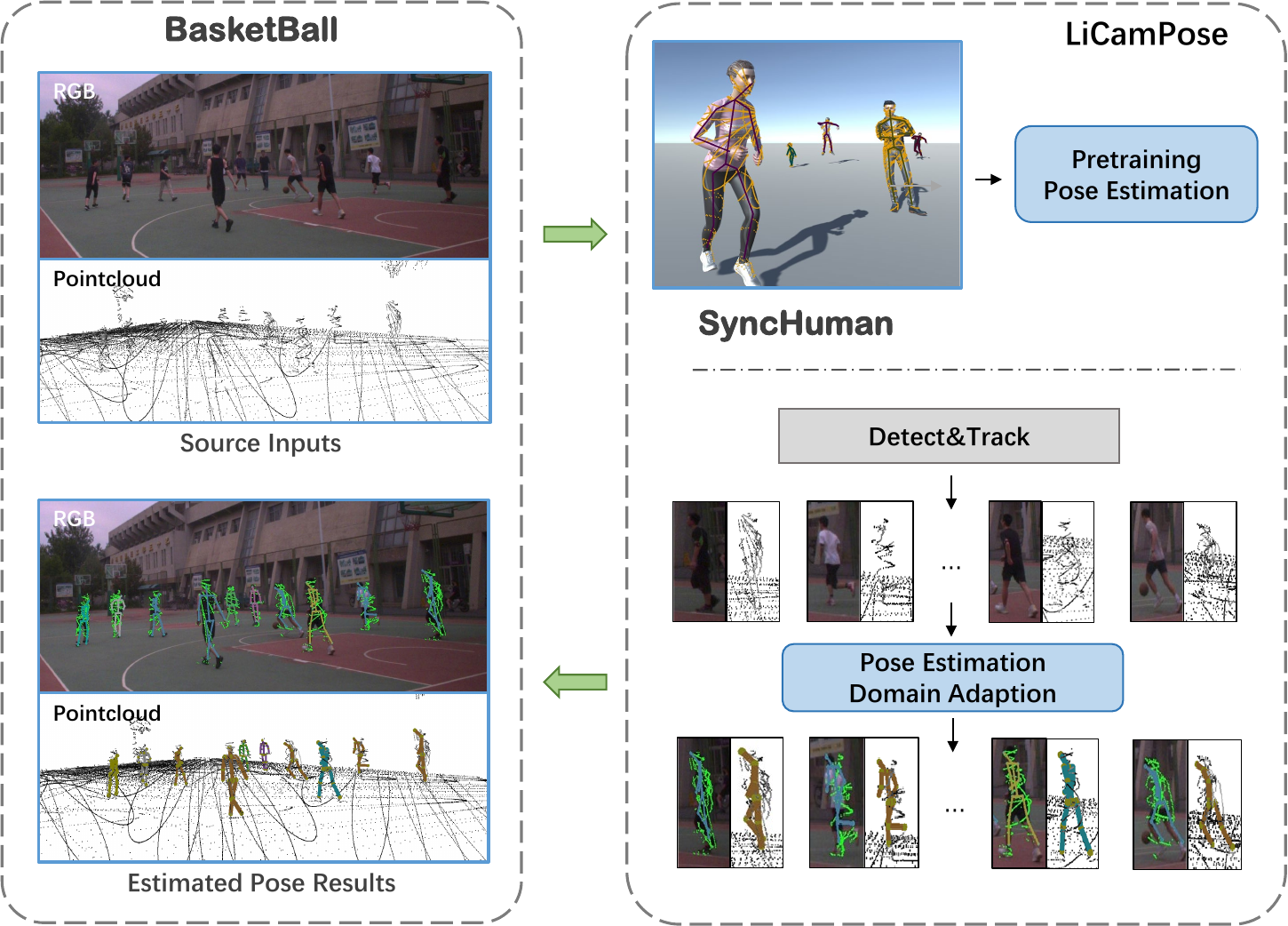}
   \caption{The LiCamPose pipeline for extracting 3D poses, as exemplified by the BasketBall dataset, involves pretraining on synthetic data from SyncHuman, followed by detecting and tracking individuals, and finally using unsupervised domain adaptation to estimate poses.}
   \label{fig:overview}
   \vspace{-0.5cm}
\end{figure}

% replace BasketBall with HoopsFusion
%  In recent years, some works have published their datasets for more challenging settings. MVOR \cite{mvor} is a multi-view RGBD dataset in operating room environment which contains many occlusions.
%  PedX \cite{pedx} and LiDARHuman26M \cite{lidarcap} are two datasets set outdoors in large scene with cameras and Velodyne LiDAR sensors. 

Fusing LiDAR point clouds with RGB camera information has been demonstrated to significantly enhance object detection and tracking performance \cite{bevfusion, yang2023bevformer}, leveraging their complementary nature.
LiDAR sensor can obtain precise but quite sparse 3D measurements over long range, while camera can capture images of high resolution but lacks depth perception. Zhang et al. \cite{zhang2022flexible} introduced a LiDAR-Camera capturing system capable of simultaneously providing texture and depth information over a wide coverage range. This advancement enables the capture of complex human activities, such as basketball games, using a minimal number of sensors, which is advantageous for system setup and cost reduction. However, their application has thus far been limited to human detection and tracking. In recent years, some researchers \cite{HPERL, cong2023weakly} have aimed to improve human pose estimation performance by temporally fusing these two modalities from a single view. Multi-view information fusion \cite{2020voxelpose,2022faster,2021tessetrack} is the most effective approach to address the inaccuracy of pose estimation caused by occlusion in single-view methods. It is essential to consider integrating the information from multiple modalities and multiple views, and fusing them into a cohesive representation which can significantly facilitate the accuracy of 3D human pose estimation.

In this paper, we introduce LiCamPose (Figure \ref{fig:overview}), a 3D human pose estimation pipeline that utilizes multi-view single-timestamp LiDAR-Camera inputs. We unify different modalities into a volumetric space that reserve each modality's geometric characteristics which facilitates better realization of single-person 3D pose estimation. Researchers \cite{2019learnable,2020voxelpose,2021tessetrack,2022faster} have shown the effectiveness of voxel-based method. Volumetric architecture can naturally model a space's geometric characteristics. Besides, it is straightforward to either map the point cloud information or back-project the 2D information into the 3D volumetric representation. 
% So it can unify different modalities into a volumetric space that reserve each modality's geometric characteristics which facilitates better realization of single-person 3D pose estimation.
Regarding how to achieve multi-person detection, some pointcloud-based methods \cite{votenet, pointpillars, bevfusion} illustrates that they can detect objects accurately with the assist of point cloud. Hence, LiCamPose combines human detection and voxel-based 3D human pose estimation in a top-down manner.

%  Manually annotating or capturing 3D poses for multiple individuals in large scenes is challenging, expensive, and hard to ensure generalizability to new scenarios. 
%  To achieve better results on various scenarios without pose labels, we adopt the approach of pretraining on a 
%  synthetic dataset first and then unsupervised domain adaption training on the target dataset.
%  We develop a synthetic dataset generator named SyncHuman for simulating data of pretraining. 
%  It can generate a large amount of synthetic data with multi-modal information, and we 
%  can change LiDARs' and cameras' properties in the scene to meet practical needs. Furthermore, the actions of avatars are diverse, as they can be 
%  obtained from existing action files or pose annotation files sourced from public motion capture datasets like AMASS \cite{amass}. 
%  Accordingly, the synthetic data has accurate 3D pose groundtruth for a group of people with complex actions.
Manually annotating or capturing 3D poses for multiple individuals in large scenes is challenging, time-consuming, and difficult. To achieve better results across diverse scenarios without relying on pose labels, we adopt a two-stage approach: first, pretraining on a synthetic dataset, and then performing unsupervised domain adaptation training on the target dataset. We have developed a synthetic dataset generator named SyncHuman for pretraining. SyncHuman can produce a large volume of synthetic data with multi-modal information, allowing us to adjust LiDAR and camera settings in scenes to meet practical requirements. Moreover, avatar actions are varied and sourced from existing action files or pose annotation files from public motion capture datasets like AMASS \cite{amass}. Consequently, the synthetic data includes accurate 3D pose ground truth for groups of people engaging in complex actions.

To bridge the gap between synthetic data and real-world scenarios, we adopt unsupervised domain adaptation training and propose an efficient strategy that includes entropy-guided pseudo 3D pose supervision, pseudo 2D pose supervision, and constraints based on human pose priors. Using volumetric representation, our approach yields 3D human joint heatmaps, with each channel representing a joint's probability distribution. We calculate each channel's entropy to gauge confidence in the corresponding joints, filtering out implausible poses and deriving pseudo 3D pose labels during unsupervised domain adaptation. We utilize off-the-shelf 2D pose estimation methods \cite{vitpose,openpose,hrnet,alphapose} to generate pseudo 2D pose labels for supervision. Additionally, we introduce an intuitive human prior loss to enforce coherence in the predicted 3D poses. By integrating these methodologies, we develop a 3D human pose estimation algorithm that leverages multi-view point cloud and RGB data without requiring annotations on the target dataset. To further validate LiCamPose, we built a four-view LiDAR-Camera system to capture a basketball game, creating the BasketBall dataset for qualitative evaluation of our method.

We summarize the contributions of this paper as follows:
\begin{itemize}
   \item We propose LiCamPose, a simple and effective pipeline for fusing multi-view, sparse point cloud and RGB information to estimate 3D poses of multiple individuals from single timestamp.
   \item We developed SyncHuman, a generator for synthetic data under various camera and LiDAR settings. Additionally, we created a four-view LiDAR-Camera system to capture real data from a basketball game, resulting in the BasketBall dataset.
   \item We propose a training strategy that avoids manual annotations by pretraining on synthetic data from SyncHuman and using unsupervised domain adaptation on the target dataset. 
  %  \item We proposed an unsupervised domain adaption training strategy to adapt the pretrained model to the real-world dataset.
\end{itemize}

%%% related work, 不整花里胡哨，直接从贡献点出发。第一部分3d姿态估计架构方法，纯多视角图像，多视角点云，以及融合的方法。
%%% 第二部分，仿真数据库大pk， 介绍几种仿真数据的优劣性，表格对比，关键强调是否提供数据产生工具
%%% 第三部分，自监督迁移学习，介绍几种方法，强调本方法，应用于多视角，简单，且解释性较好。 （这里放几种自监督弱监督迁移的方法）

\section{Related Works} 
In this section, we conduct the literature review according to the three contribution points we proposed.
\subsection{3D Human Pose Estimation}
\noindent
\textbf{Image-based.}
% Basic 3D pose estimation method is in two stage that estimates the 2D pose first and then lifts it into 3D space  
% \cite{2019fast,2019learnable,2020voxelpose,2021multi,2021graph,2021tessetrack}. As to multi-person setting, some methods \cite{2019fast,2021multi} match pedestrians from 
% different views and then locate them through 2D pose similarity. But they are not robust to inaccurate 2D pose results. Zhang et al. \cite{2021direct} 
% directly utilizes the 2D images as inputs and regresses the 3D pose by transformer architecture. However, their training process is time-consuming. Some voxel-based methods \cite{2020voxelpose, 2021tessetrack} locate each person in a 3D volumetric space and estimate 3D poses. 
% Such voxel-based methods greatly improve the precision of 3D human pose estimation. However, they are not suitable for the large scene because of 
% large computational cost when detecting people.
The basic 3D pose estimation method typically follows a two-stage process: first estimating the 2D pose and then lifting it into 3D space \cite{2019fast,2019learnable,2020voxelpose,2021multi,2021graph,2021tessetrack}. For multi-person settings, some methods \cite{2019fast,2021multi} match pedestrians from different views and locate them through 2D pose similarity. However, these methods are not robust to inaccurate 2D pose results. Zhang et al. \cite{2021direct} directly use 2D images as inputs and regress the 3D pose with a transformer architecture, but their training process is time-consuming. Voxel-based methods \cite{2020voxelpose,2021tessetrack} locate each person in a 3D volumetric space and estimate 3D poses, significantly improving the precision of 3D human pose estimation. However, these voxel-based methods are not suitable for large scenes due to their high computational cost when detecting people. LiCamPose utilizes pointcloud-based method to locate and track pedestrians, and then employs voxel-based pose estimation for each individual.

\noindent
\textbf{Pointcloud-based.}
Initially, several 3D pose estimation methods were based on single-view depth maps \cite{garau2021deca,haque2016towards,guo2017towards,martinez2020residual}, treating the depth map as 2D information with depth values. Conversely, some methods back-project the depth map into 3D space as a dense 3D point cloud and utilize PointNet networks \cite{zhang2021sequential,bekhtaoui2020view}. Moon et al. \cite{moon2018v2v} proposed a single-person pose estimation approach that treats the depth map as a point cloud and fills it into a volumetric space. Bekhtaoui et al. \cite{bekhtaoui2020view} employed a PointNet-based approach to detect and estimate 3D human pose. More recently, Li et al. \cite{lidarcap} used sparse LiDAR-scanned point clouds to estimate the 3D pose of a single person. Zhang et al. \cite{zhang2024neighborhood} enhanced performance by incorporating the point cloud surrounding the person as a neighborhood context. However, temporal information is essential for the effectiveness of both approaches. Sparse point clouds from single timestamp do not provide sufficient information for accurate 3D pose estimation. Multi-modal fusion is beneficial for precise 3D human pose perception.

\noindent
\textbf{Multi-modal based.}
Several works have been introduced for RGBD human pose estimation \cite{bashirov2021real,zheng2022multi,hansen2019fusing,gerats20223d}, demonstrating that multi-modal information not only aids in detecting individuals but also ensures the accuracy of 3D human pose estimation.
And recently, researchers \cite{HPERL, cong2023weakly} tried to fuse the RGB and point cloud temporal information from singe view to improve the pose estimation accuracy. However, they rely on temporal information for supplementation, and when a single view is occluded for an extended period, the methods fail to make accurate predictions. Regarding point cloud and RGB information fusion methods, some approaches \cite{zheng2022multi,bekhtaoui2020view,hansen2019fusing} fuse these modalities at the point level by attaching extracted 2D features to each 3D point. A few methods \cite{piergiovanni20214d,liang2018deep} employ a feature-level fusion strategy. Nevertheless, these methods do not integrate the different modalities in a feature space that preserves each modality's spatial properties. In contrast, LiCamPose employs a volumetric representation to integrate point cloud and RGB data from multiple views within a single timestamp, preserving the spatial properties of the real-world environment and mitigating the impact of occlusion.

\begin{table}[t]\scriptsize
   \caption{Comparison of synthetic datasets related to human.}
   \label{tab:syncdataset}
   \centering
   \resizebox{ \linewidth}{!}{   
\begin{tabular}{l p{8pt}p{8pt}p{8pt}p{8pt}p{8pt} p{8pt}p{8pt}p{8pt}}
\toprule
\multicolumn{1}{c}{\multirow{2}{*}{\begin{tabular}[c]{@{}c@{}}Synthetic\\ Dataset\end{tabular}}} & \multicolumn{5}{c}{Scene setup}                                                                                                                                                                                                                                                                                                                         & \multicolumn{2}{c}{Labels}                                                                                                                   \\ \cmidrule(lr){2-3} \cmidrule(lr){4-6} \cmidrule(lr){7-8}  %\cline{2-8} 
\multicolumn{1}{c}{}                                                                             & \multicolumn{1}{c}{\begin{tabular}[c]{@{}c@{}}Multi-\\ View\end{tabular}} & \multicolumn{1}{c}{\begin{tabular}[c]{@{}c@{}}Multi-\\ Person\end{tabular}} & \begin{tabular}[c]{@{}l@{}}RGB\\ Image\end{tabular} & \begin{tabular}[c]{@{}l@{}}Depth\\ Image\end{tabular} & \multicolumn{1}{c}{\begin{tabular}[c]{@{}c@{}}LiDAR\\ Pointcloud\end{tabular}} & \multicolumn{1}{c}{\begin{tabular}[c]{@{}c@{}}2D\\ Pose\end{tabular}} & \multicolumn{1}{c}{\begin{tabular}[c]{@{}c@{}}3D\\ Pose\end{tabular}}\\ \midrule %\cline{1-8} 
CAPE \cite{cape}                                                                                             & \multicolumn{1}{c}{\usym{2717}}                                                          &  \multicolumn{1}{c}{\usym{2717}}                                                        &  \multicolumn{1}{c}{\usym{2717}}                                                   &  \multicolumn{1}{c}{\usym{2717}}                                                      & \multicolumn{1}{c}{\usym{2717}}                                                                                &   \multicolumn{1}{c}{\usym{2717}}                                                                     & \multicolumn{1}{c}{\usym{2713}}                                                 \\
SURREAL \cite{surreal}                                                                                             & \multicolumn{1}{c}{\usym{2717}}                                                          &  \multicolumn{1}{c}{\usym{2717}}                                                        &  \multicolumn{1}{c}{\usym{2713}}                                                   &  \multicolumn{1}{c}{\usym{2713}}                                                      & \multicolumn{1}{c}{\usym{2717}}                                                                                &   \multicolumn{1}{c}{\usym{2713}}                                                                     & \multicolumn{1}{c}{\usym{2713}}                                                 \\
PSP \cite{peoplesanspeople}                                                                                              & \multicolumn{1}{c}{\usym{2717}}                                                          &  \multicolumn{1}{c}{\usym{2717}}                                                        &  \multicolumn{1}{c}{\usym{2713}}                                                   &  \multicolumn{1}{c}{\usym{2717}}                                                      & \multicolumn{1}{c}{\usym{2717}}                                                                                &   \multicolumn{1}{c}{\usym{2713}}                                                                     & \multicolumn{1}{c}{\usym{2717}}                                                 \\
AGORA \cite{AGORA} & \multicolumn{1}{c}{\usym{2717}} & \multicolumn{1}{c}{\usym{2713}} & \multicolumn{1}{c}{\usym{2713}} & \multicolumn{1}{c}{\usym{2717}} & \multicolumn{1}{c}{\usym{2717}} & \multicolumn{1}{c}{\usym{2713}} & \multicolumn{1}{c}{\usym{2713}}\\
BEDLAM \cite{BEDLAM} & \multicolumn{1}{c}{\usym{2717}} & \multicolumn{1}{c}{\usym{2713}} & \multicolumn{1}{c}{\usym{2713}} & \multicolumn{1}{c}{\usym{2717}} & \multicolumn{1}{c}{\usym{2717}} & \multicolumn{1}{c}{\usym{2717}} & \multicolumn{1}{c}{\usym{2713}}\\
BlendMinic3D \cite{BlendMimic3D} & \multicolumn{1}{c}{\usym{2713}} & \multicolumn{1}{c}{\usym{2713}} & \multicolumn{1}{c}{\usym{2713}} & \multicolumn{1}{c}{\usym{2717}} & \multicolumn{1}{c}{\usym{2717}} & \multicolumn{1}{c}{\usym{2713}} & \multicolumn{1}{c}{\usym{2713}}\\
CALAR \cite{carla}                                                                                            & \multicolumn{1}{c}{\usym{2713}}                                                          &  \multicolumn{1}{c}{\usym{2713}}                                                        &  \multicolumn{1}{c}{\usym{2713}}                                                   &  \multicolumn{1}{c}{\usym{2713}}                                                      & \multicolumn{1}{c}{\usym{2713}}                                                                                &   \multicolumn{1}{c}{\usym{2717}}                                                                     & \multicolumn{1}{c}{\usym{2717}}                                                 \\
\textit{Ours}                                                                                     & \multicolumn{1}{c}{\usym{2713}}                                                          &  \multicolumn{1}{c}{\usym{2713}}                                                        &  \multicolumn{1}{c}{\usym{2713}}                                                   &  \multicolumn{1}{c}{\usym{2713}}                                                      & \multicolumn{1}{c}{\usym{2713}}                                                                                &   \multicolumn{1}{c}{\usym{2713}}                                                                     & \multicolumn{1}{c}{\usym{2713}}                                                 \\ \bottomrule %\cline{1-8}
\end{tabular}}
\end{table}

\subsection{Synthetic Dataset Generation}
% Manually annotating 3D human pose is very tough. Therefore, synthetic dataset 
% is beneficial and helpful for pretraining model. There are many synthetic datasets for 3D human pose estimation \cite{surreal,cape,peoplesanspeople}. 
% However, these datasets are just single RGB view with random background. Besides, they do not consider the rational interaction between  
% avatars and the background. Dosovitskiy et al. \cite{carla} provides a large-scale synthetic dataset for autonomous driving. However, it does not provide the 3D human pose labels, and the actions of characters are monotonous. 
Manually annotating 3D human poses is extremely challenging. Therefore, synthetic datasets are beneficial and useful for pretraining models. Several synthetic datasets for 3D human pose estimation have been developed \cite{surreal,cape,peoplesanspeople, AGORA, BEDLAM, BlendMimic3D}. However, these datasets typically feature an RGB photo from one view with random backgrounds and do not consider the realistic interaction between avatars and their environment. Additionally, they all lack the provision of point cloud information. Dosovitskiy et al. \cite{carla} provide a large-scale synthetic dataset for autonomous driving, but it lacks 3D human pose labels, and the actions of the characters are monotonous.

We introduce our synthetic dataset generator, SyncHuman, which can produce data with greater richness and diversity (Table \ref{tab:syncdataset}). Furthermore, we will release the generation tool, allowing researchers to adjust settings according to their needs via our provided APIs

\subsection{Unsupervised Domain Adaption Training}
Numerous works have focused on unsupervised or domain adaptation methods for single-view 3D human pose estimation. Kocabas et al. \cite{2019self} use multi-view geometry to supervise single-view predictions. Several methods \cite{srivastav2020self,kundu2020unsupervised} employ a teacher-student framework for domain adaptation. Some approaches \cite{doersch2019sim2real,zhang2019unsupervised} utilize optical flow or depth as inputs, which are less affected by domain shifts compared to RGB. Bigalke et al. \cite{bigalke2022domain} incorporate human prior loss based on human anatomy, while Kundu et al. \cite{2022uncertainty} define the uncertainty of predictions and control this uncertainty during training. For multi-view 3D human pose generation, some works use non-deeplearning methods \cite{3dps, remelli2020lightweight}. 3DPS \cite{3dps} and basic triangulation achieve 3D poses with significant computational complexity or inaccuracy. Remelli et al. \cite{remelli2020lightweight} proposed an efficient direct linear transformation (DLT) method that quickly produces relatively accurate 3D results.

Our approach utilizes multi-view 2D pose heatmaps derived from RGB data alongside point clouds, ensuring minimal impact from domain variations. We utilize information entropy to help select reliable results as pseudo 3D labels. Additionally, we incorporate a human prior loss to ensure the plausibility of 3D poses.

%% 方法论撰写，由于在网络创新点的成果较少，主要以简短的流程介绍，以及自己想到的不同点为主。在定位的时候提出，纯rgb的方法较点云方法的劣势
%% 简要介绍LiCamPose方法的组成部分，由点云检测场景中人，并投影2d ，2d predict, 

\begin{figure*}[!ht]
   \centering
   \includegraphics[width=0.9\linewidth]{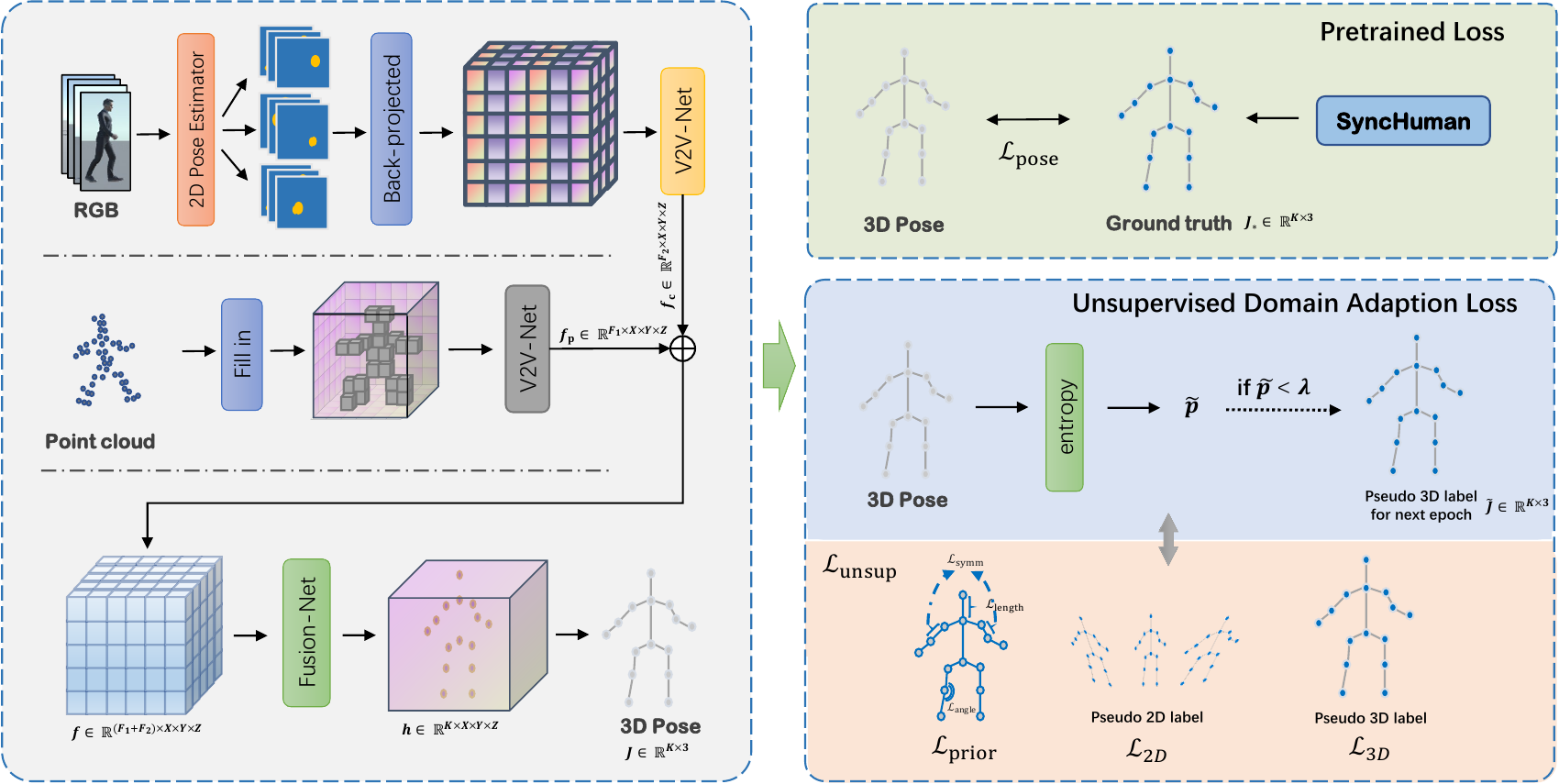}
   \caption{The detailed structure of LiCamPose in 3D human pose estimation and its corresponding losses calculations.} 
   \label{fig:pointvoxel}
   \vspace{-0.5cm}
\end{figure*}   

\vspace{-0.2cm} % TODO:
\section{Methodology}
We introduce LiCamPose from two perspectives: 3D human pose estimation and unsupervised domain adaptation (Figure \ref{fig:pointvoxel}). Section \ref{pointvoxel} details the top-down approach for 3D human pose estimation. Section \ref{unsup} explains how we generate the synthetic dataset to support the training of the pose estimation model using point cloud and RGB data, and how we utilize information entropy and human prior loss to achieve domain adaptation.
\subsection{3D Human Pose Estimation} \label{pointvoxel}
% Similar to \cite{2020voxelpose, 2019fast,2021tessetrack}, we adopt the top-down manner to estimate 3D poses. In our case, we adopt PointPillars \cite{pointpillars} to detect the 3D bounding box of each person. After obtaining the bounding box of each person, we can get the 2D bounding box in each view for 2D pose estimation, and extract their point cloud. At first, we can define a volumetric space centered at the poincloud's centroid whose size is consistent with detected bounding boxes', and discretize it into an $X \times Y \times Z$ resolution. So that we can fill each voxel according to each pointcloud's coordinate. In our case, we set value $1$ to the voxel containing poincloud. Then we can get the pointcloud-related feature ($F_1 \times X \times Y \times Z$) via 3D convolution backbone V2V-Net \cite{moon2018v2v}.
Similar to \cite{2020voxelpose, 2019fast,2021tessetrack}, we adopt a top-down approach to estimate 3D poses. Specifically, we use PointPillars \cite{pointpillars} to detect the 3D bounding box of each person. Once we obtain the 3D bounding box, we derive the 2D bounding box in each view for 2D pose estimation and extract the corresponding point cloud. Initially, we define a volumetric space centered at the point cloud's centroid, with a size consistent with the detected bounding boxes, and discretize it into an $X \times Y \times Z$ resolution. We fill each voxel based on the coordinates of the point cloud, assigning a value of $1$ to voxels containing points from the point cloud. This allows us to obtain the pointcloud-related feature map $f_\text{p} \in \mathbb{R}^{F_1 \times X \times Y \times Z}$ using the 3D convolutional backbone V2V-Net \cite{moon2018v2v}.

\begin{figure*}[!t]
   \centering  
   \includegraphics[width=.8\linewidth]{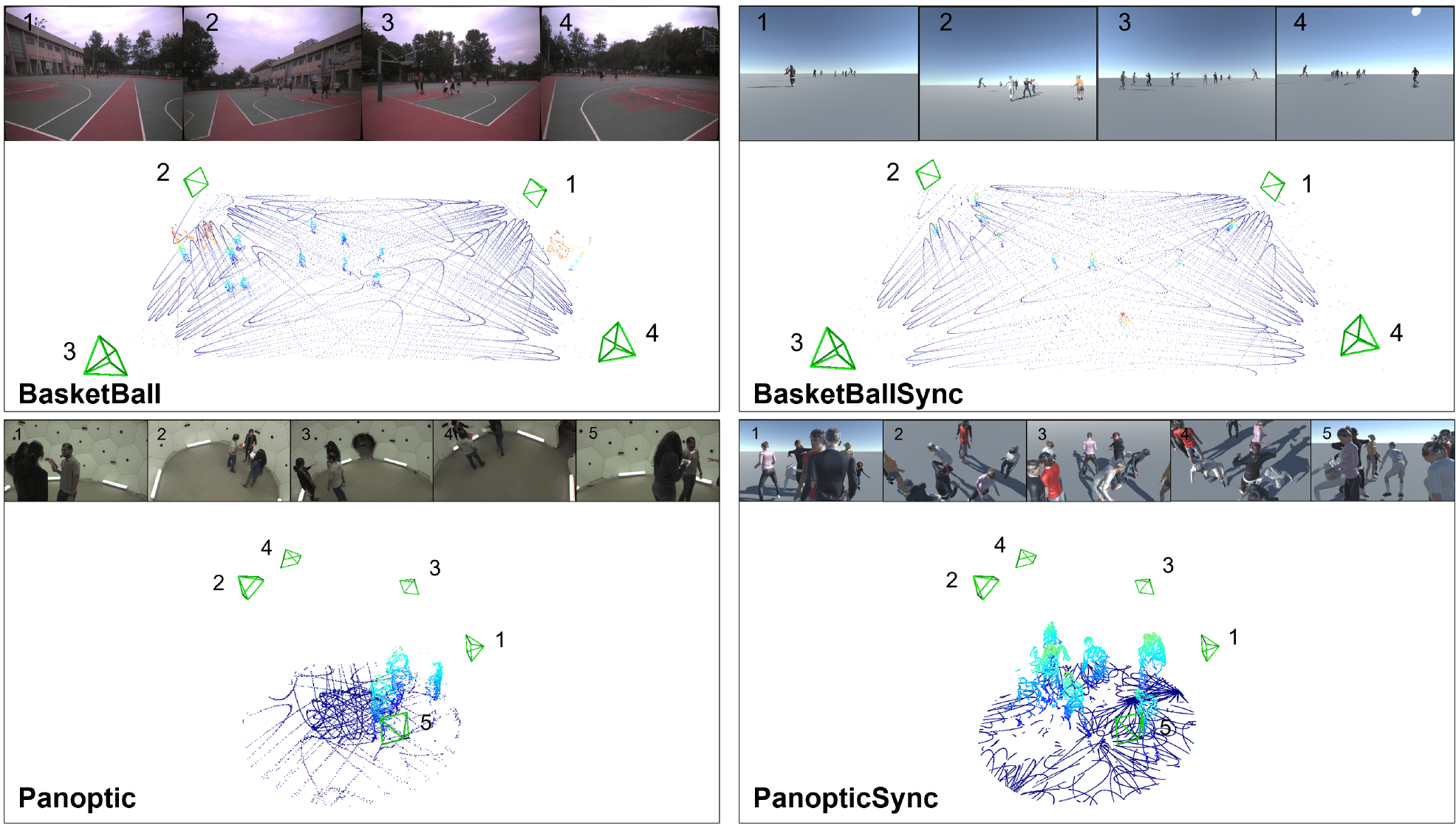}
   \caption{Real datasets and the corresponding synthetic datasets generated by SyncHuman.}
   \label{fig:dataset}
   \vspace{-0.3cm}
\end{figure*}

\begin{figure}[!t]
   \centering  
   \includegraphics[width=.8\linewidth]{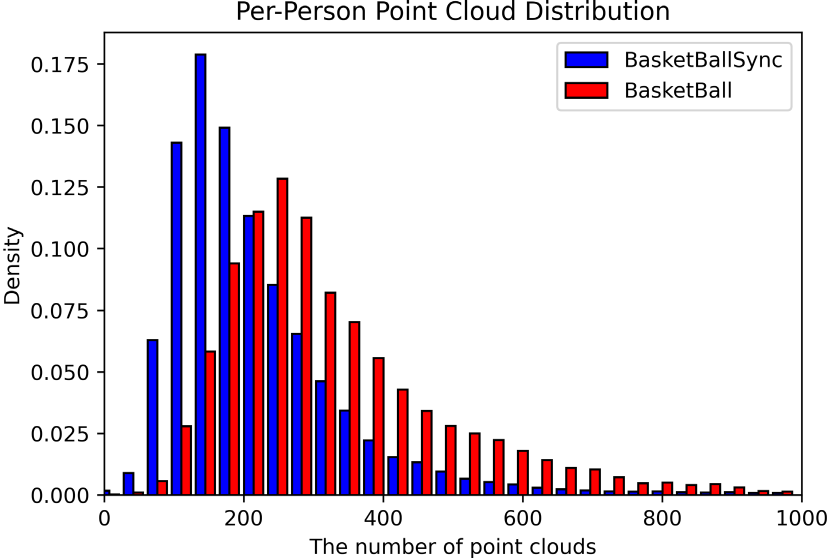}
   \caption{Statistic analysis of per-person point cloud counts between BaseketBall and BaseketBallSync.}
   \label{fig:dataset_comp}
   \vspace{-0.5cm}
\end{figure}

% Thanks to the development of 2D pose estimation methods \cite{alphapose,vitpose}, robust 2D pose results can be predicted using off-the-shelf methods. Hence, we utilize AlphaPose \cite{alphapose} to get each view's 2D pose heatmap from RGB information. We backproject multi-view heatmaps into the volumetric space according to the setting of each camera and use V2V-Net to extract the rgb-related features ($F_2 \times X \times Y \times Z$) in 3D space. Then, we concatenate the two modal features in the same 3D space for modality fusion and get the final 3D human pose heatmap via a Fusion-Net. In order to mitigate quantization error, we adopt \textit{Soft-argmax} to calculate each joints' 3D coordinate $J^k$ through 3D heatmaps and minimize the $L_1$ loss with the ground truth $J_*^k$: 
With the development of 2D pose estimation methods \cite{alphapose,vitpose}, robust 2D pose results can be predicted using off-the-shelf techniques. We use these methods to obtain 2D pose heatmaps from RGB information of each view. These multi-view heatmaps are then back-projected into the volumetric space according to each camera's settings, and V2V-Net is used to extract the RGB-related features $f_\text{c} \in \mathbb{R}^{F_2 \times X \times Y \times Z}$. We concatenate the features from both modalities in the same 3D feature space to get fusion feature $f$ and obtain the final 3D human pose heatmap $h$ via Fusion-Net, which has the same structure as V2V-Net. To mitigate quantization error, we adopt \textit{Soft-argmax} to calculate each joint's 3D coordinate $J^k$ from the 3D heatmaps and minimize the $L_1$ loss with the ground truth $J_*^k$:
\begin{equation}
\begin{aligned}
   \label{poseloss}
   \mathcal{L}_{\text{pose}} = \sum_{k=1}^{K} \left \Vert J^k - J_*^k \right \Vert_1,
\end{aligned}
\end{equation}
where $K$ is the number of joints and $J^k$ is the predicted joint, $J^k, J_*^k \in \mathbb{R}^3$.

\subsection{Unsupervised Domain Adaptation} \label{unsup}
% Due to the lack of annotated multi-view LiDAR-Camera 3D human pose dataset, we choose to generate a synthetic dataset to assist training. SyncHuman we designed can create 3D dynamic scenes of multiple persons with accurate labels. Besides, we develop an efficient unsupervised domain adaption means by designing a loss function to transfer the pretrained model from synthetic dataset to real-world dataset. 
Due to the lack of annotated multi-view LiDAR-Camera 3D human pose datasets, we opted to generate a synthetic dataset to assist with training. Our designed SyncHuman can create 3D dynamic scenes with multiple persons and accurate labels. Additionally, we developed an efficient unsupervised domain adaptation method by designing a loss function to transfer the pretrained model from the synthetic dataset to a real-world dataset. 
\subsubsection{SyncHuman Generator} \label{sync}
% We develop the synthetic system based on Unity Engine. Due to its flexibility and productivity, we can formulate a scene with a specific size and background as per our specifications. Concerning sensors' aspect, we implement Unity's built-in camera to obtain RGB images, and extract the depth information from the GPU's depth buffer with custom shaders used in the rendering pipeline.  Furthermore, we can attain the colored point cloud by sampling the depth and RGB images. For the sake of imitating the scanning process of LiDAR, we need to sample points along to a scanning function of time. Currently, we consider Livox Mid-40 LiDAR only, since its price is sufficiently low for sports or surveillance applications.
We developed the synthetic system based on the Unity Engine. Its flexibility and productivity allow us to create scenes with specific sizes and backgrounds according to our requirements. For sensors, we use Unity's built-in camera to obtain RGB images and extract depth information from the GPU's depth buffer using custom shaders in the rendering pipeline. Additionally, we can generate colored point clouds by sampling the depth and RGB images. To simulate the LiDAR scanning process, we sample points according to a time-based scanning function. We currently focus on the Livox Mid-40 LiDAR due to its affordability for sports or surveillance applications.
We use the following function to simulate its scanning process: 
\begin{equation}
\begin{aligned}
   \label{scan}
   r = \alpha \times \mathbb{\cos}(3.825\times(\theta_0+0.0017\times n)),
\end{aligned}
\end{equation}
where $n \in [0,t\times 1e5]$ and $t$ is in second. $\alpha$ is the scanning radius in pixel, and $\theta_0$ is a random initial angle. This equation is defined in polar coordinate. Final sampling points can be acquired by transforming it from polar coordinates to Cartesian coordinates and translating it to the center of the image space.

% As for the avatars, we download various human 3D models from Adobe Mixamo\footnote{https://mixamo.com}. With a view to guarantee the diversity of the generated actions, we can drive these avatars by either ready-made action files or other public datasets' keypoint annotations. Currently, we have developed the driving APIs for COCO17, COCO19 \cite{lin2014microsoft} and SMPL \cite{loper2015smpl} standard keypoint annotated inputs. Regarding the groundtruth matter, we can obtain the 3D human pose from the avatars' humanoid skeleton; 2D pose label can be acquired via projecting 3D pose into 2D view. Additionally, we can fetch the mesh vertex of each avatar and then compute the semantic segmentation label for each point cloud by considering the pose label meanwhile.     
For the avatars, we download various human 3D models from Adobe Mixamo\footnote{https://mixamo.com}. To ensure the diversity of the generated actions, these avatars can be driven by either pre-made action files or keypoint annotations from other public datasets such as AMASS \cite{amass}. We have developed driving APIs for COCO17, COCO19 \cite{lin2014microsoft}, and SMPL \cite{loper2015smpl} standard keypoint annotated inputs. Regarding ground truth data, we can obtain the 3D human pose from the avatars' humanoid skeleton; the 2D pose label can be acquired by projecting the 3D pose into 2D views. Additionally, we can fetch the mesh vertices of each avatar and compute the semantic segmentation label for each point cloud by considering the pose label simultaneously.

We can use SyncHuman to simulate any arrangement of sensors to observe a scene. As demonstrated in the experiments in our paper, using the same scene setting for both training and testing yields better transfer performance. Figure \ref{fig:dataset} compares the datasets we generated, PanopticSync and BasketBallSync, with the original datasets. Since SyncHuman allows for flexible sensor placement, the distribution of the number of point clouds scanned per person in the simulation closely approximates the real-world distribution (Figure \ref{fig:dataset_comp}).

\vspace{-0.3cm}
\subsubsection{Unsupervised Domain Adaptation} \label{self}
%% 重点介绍熵值计算，带过一点人体解剖学误差
% Similar to the 2D projection supervision utilized by other unsupervised or weakly supervised methods \cite{2019self,gholami2022self}, we directly 
% employ off-the-shelf 2D human pose estimation model to get the pseudo 2D pose label $\widetilde{J}_{\text{2D}}$. $\mathcal{L}_{\text{2D}}$ is obtained by calculating the 
% $L_2$ norm between 3D pose projection results and pseudo 2D label from each view:
Similar to the 2D projection supervision used by other unsupervised or weakly supervised methods \cite{2019self,gholami2022self}, we directly utilize an off-the-shelf 2D human pose estimation method to obtain the pseudo 2D pose label $\widetilde{J}{\text{2D}}$. The 2D loss $\mathcal{L}{\text{2D}}$ is computed by calculating the $L_2$ norm between the 3D pose projection results and the pseudo 2D label from each view:
\begin{equation}
\begin{aligned}
   \label{2dloss}
   \mathcal{L}_{\text{2D}} =  \sum_{v=1}^{V}\sum_{k=1}^{K} \left \Vert \mathcal{P}_v(J^{v,k}) - \widetilde{J}_{\text{2D}}^{v,k} \right \Vert_2,
\end{aligned}
\end{equation}
where $\mathcal{P}$ is the projection function. $V$ is the number of views. 
% $\mathcal{L}_{\text{2D}}$ is the fundamental loss for unsupervised training.
% In order to attain pseudo 3D label, we use information 
% entropy as an uncertainty index. The entropy of one keypoint prediction heatmap (a spatial probability distribution) $h^k$ is defined as:
To obtain the pseudo 3D label, we use information entropy as an uncertainty index. The entropy of a keypoint prediction heatmap (a spatial probability distribution) $h^k$ is defined as:
\begin{equation}
\begin{aligned}
   \label{entropy}
   \mathcal{H}(h^k) = -\sum_{i} h_i^k \times \log h_i^k,
\end{aligned}
\end{equation}
% where $i$ represents the voxel index. The higher the entropy is, the more uncertain the keypoint location is. Furthermore, it relates to the rationality of the keypoint location as observed in our experiments (Section \ref{entropy_ana}). 
where $i$ represents the voxel index. Higher entropy indicates greater uncertainty in the keypoint location. Moreover, as observed in our experiments (Section \ref{entropy_ana}), there is a correlation between entropy and the rationality of the keypoint location.

% To measure a person's uncertainty, we take the maximum entropy value of all keypoints. Different from Kundu et al. \cite{2022uncertainty} who train the uncertainty value, we consider it as an inartificial index. Our experiments demonstrate that the magnitude of entropy values can serve as an indicator of pose estimation quality for a specific network. Therefore, an entropy threshold $\lambda$ can be 
% set to sort out the reasonable predicted 3D human poses. In other words, we select the predicted pose $\widetilde{p}$ whose entropy value is less than $\lambda$ as the pseudo 3D pose 
% label for the next epoch's training, and it can be updated after each training epoch. Therefore, the 3D pseudo pose loss can be obtained by:
To measure a person's uncertainty, we use the maximum entropy value of all keypoints. Unlike Kundu et al. \cite{2022uncertainty}, who train a model to learn the uncertainty values, we directly calculate and use entropy as an inherent index. Our experiments demonstrate that the magnitude of entropy values can serve as an indicator of pose estimation quality for a specific network. Consequently, an entropy threshold $\lambda$ can be set to filter out the reasonably predicted 3D human poses. Specifically, we select the predicted pose $\widetilde{p}$ whose entropy value is less than $\lambda$ as the pseudo 3D pose label for the next epoch's training, allowing it to be updated after each training epoch. Therefore, the 3D pseudo pose loss can be obtained by:
\begin{equation}
\begin{aligned}
   \label{3dloss}
   \mathcal{L}_{\text{3D}} = \sum_{k=1}^{K} \left \Vert J^k - \widetilde{J}^k \right \Vert_1,
\end{aligned}
\end{equation}
where $J^k$ represent the predicted joints and $\widetilde{J}^k$ is the 3D joint of the selected pseudo 3D pose label $\widetilde{p}$.

% To further ensure the anatomical plausibility of the pose, we design a human prior loss $\mathcal{L}_{\text{prior}}$\footnote{Please refer to Suppl. for details.}. Specifically, we formulate three losses to penalize asymmetric limb lengths $\mathcal{L}_{\text{symm}}$, implausible joint angles $\mathcal{L}_{\text{angle}}$, and implausible bone lengths $\mathcal{L}_{\text{length}}$. 
To further ensure the anatomical plausibility of the pose, we design a human prior loss $\mathcal{L}_{\text{prior}}$\footnote{Please refer to the Supplementary Material for details.}. Specifically, we formulate three losses to penalize asymmetric limb lengths ($\mathcal{L}_{\text{symm}}$), implausible joint angles ($\mathcal{L}_{\text{angle}}$), and implausible bone lengths ($\mathcal{L}_{\text{length}}$).

In summary, the final loss function is defined as:
\begin{equation}
\begin{aligned}
   \label{loss}
   \mathcal{L}_{\text{unsup}} =\omega_\text{1} \mathcal{L}_{\text{2D}} + \omega_\text{2} \mathbf{1}(\widetilde{p}<\lambda)\mathcal{L}_{\text{3D}} + \omega_\text{3} \mathcal{L}_{\text{prior}},
\end{aligned}
\end{equation}
where $\omega_\text{1}$, $\omega_\text{2}$ and $\omega_\text{3}$ are the weights of each loss, and $\mathbf{1}(\cdot)$ represents the indicator function. 

% \begin{figure}
% \centering
% \includegraphics[width=1\linewidth]{figs/modality.pdf}
% \caption{Qualitative illustration on the BasketBall dataset from different input modalities. The first row shows 2D pose estimations (missing where not estimable) and point clouds. The second row displays results from using only RGB input, with 2D poses projected from the estimated 3D poses. The third row presents results from using both RGB and point cloud inputs.}
% \label{fig:basketball}
% \end{figure}

% \begin{figure*}[!ht]
% \centering
% \includegraphics[width=0.9\linewidth]{figs/ablation_updated.png} 
% \caption{Qualitative visualization on BasketBall about different unsupervised training losses. ``Baseline'' uses only pseudo 2D pose supervision. ``Entropy'' indicates the addition of entropy-selected pseudo 3D pose supervision. ``Prior'' denotes the incorporation of human prior loss.}
% \label{fig:basketball2}
% \end{figure*}

% 实验部分的叙述往详细走，不要太简略 （前文内容可做删减） 1. 检测准确度 2. 证明point voxel方法有效性，多库监督学习 3. 无监督层面，3.1 仿真数据库的设定对迁移的影响 3.2 无监督迁移loss部分的多库消融实验（定性定量）
% 3.3 无监督迁移的结果定性分析，与mvor等库的监督方法对比，甚至可以与真值进行定性对比。 4. More analysis，熵值的正相关性与分布 （柱状图以及分布效果【人为根据xxx标准统计结果】）
\section{Experiments}
In this section, we introduce all datasets and evaluation metrics used in our experiments. We analyze the performance of our LiCamPose pipeline in supervised and unsupervised manner on different datasets. Additionally, we examine the impact of different configurations on the results in the context of unsupervised domain adaptation.

\subsection{Implementation details.} 
We use V2V-Net \cite{moon2018v2v} as the feature extractor for point cloud and RGB information. The detailed design of V2V-Net follows the PRN model \cite{2020voxelpose}. The resolution of the voxel grid is set to \(64 \times 64 \times 64\) within a \(2 \text{m} \times 2 \text{m} \times 2 \text{m}\) space. The weights are set as \(\omega_{1} = 0.02\), \(\omega_{2} = 1\), \(\omega_{3} = 10\), and \(\lambda = 6\). We train the voxel-based pose estimation network on an NVIDIA GeForce RTX-3090 with a batch size of eight. The learning rate is set to 0.001, and the optimizer used is Adam \cite{kingma2014adam}.

\subsection{Datasets and Metrics.} 
\vspace{-0.2cm}
We use the following datasets in our experiments:\\
\noindent \textbf{I. CMU Panoptic Studio \cite{panoptic}.} 
The setup is indoors with a valid scene range of approximately \(5 \text{m} \times 5 \text{m}\). For multi-modal inputs, we select subsets\footnote{``160422\_ultimatum1", ``160224\_haggling1", ``160226\_haggling1", ``161202\_haggling1", ``160906\_ian1", ``160906\_ian2", ``160906\_ian3", and ``160906\_band1" for training; ``160906\_pizza1", ``160422\_haggling1", ``160906\_ian5", ``160906\_band2" for testing.} from \cite{2021multi} that include depth information. We unify the depth data from Kinect 1 to 5 and process these depth maps using equation \ref{scan} to obtain a sparse point cloud, similar to Mid-40 Livox LiDAR scanning. For 2D pose estimation, we use the results from \cite{2021multi}, predicted with HRNet \cite{hrnet}.\\
\noindent \textbf{II. MVOR \cite{mvor}.} 
The dataset was collected indoors in an operating room. We use three sampling functions (equation \ref{scan}) centered at three trisection points of the image width to simulate Mid-100 Livox LiDAR scanning. For 2D pose estimation, we use results from \cite{mvor}, predicted using AlphaPose \cite{alphapose}. Due to a semantic keypoints' definition gap between MVOR (ten keypoints) and our COCO17 standard, we only perform qualitative analysis on MVOR. Data from "day1, day2, day3" are used for training, and "day4" for testing.\\
\noindent \textbf{III. BasketBall.} 
The dataset was collected outdoors, covering a valid scene range of approximately \(35 \text{m} \times 17 \text{m}\), recording a basketball match. We gathered this real-world dataset with four-view RGB-poincloud information. The point clouds were collected using four Mid-100 Livox LiDARs, and the 2D pose estimation results were predicted by VitPose \cite{vitpose}. The dataset comprises two segments, each containing two thousand frames. It includes manually marked detection and tracking ground truth for each frame but lacks 3D keypoints ground truth. Therefore, it is primarily used for qualitative analysis.\\ 
\noindent \textbf{IV. BasketBallSync.} 
This synthetic dataset was generated using SyncHuman with the same sensor configuration and settings as BasketBall dataset. The point cloud is obtained by simulating the scan pattern of Mid-100 Livox LiDARs. It contains ten avatars downloaded from Mixamo\footnote{Character 2, 6, 11, 22, 23, 31, 37, 38, 42 and a kid model named ``Gregory''.} performing random actions, with the first eight used for training and the remaining two for testing, across all 3336 frames at 10Hz. We utilize the pose data from the ``CMU'' track of AMASS\cite{amass} to animate the avatars. The 2D human poses are predicted using VitPose \cite{vitpose}. \\
\noindent \textbf{Evaluation metrics.} 
To evaluate the 3D human pose, we calculate the mean per-joint position error (MPJPE) and the error after Procrustes Alignment (PA-MPJPE), both in millimeters \cite{gower1975generalized}. For fairness, we compare our results to methods using single-timestamp data inputs similar to ours.

\begin{figure}[!t]
   \centering
   \includegraphics[width=0.9\linewidth]{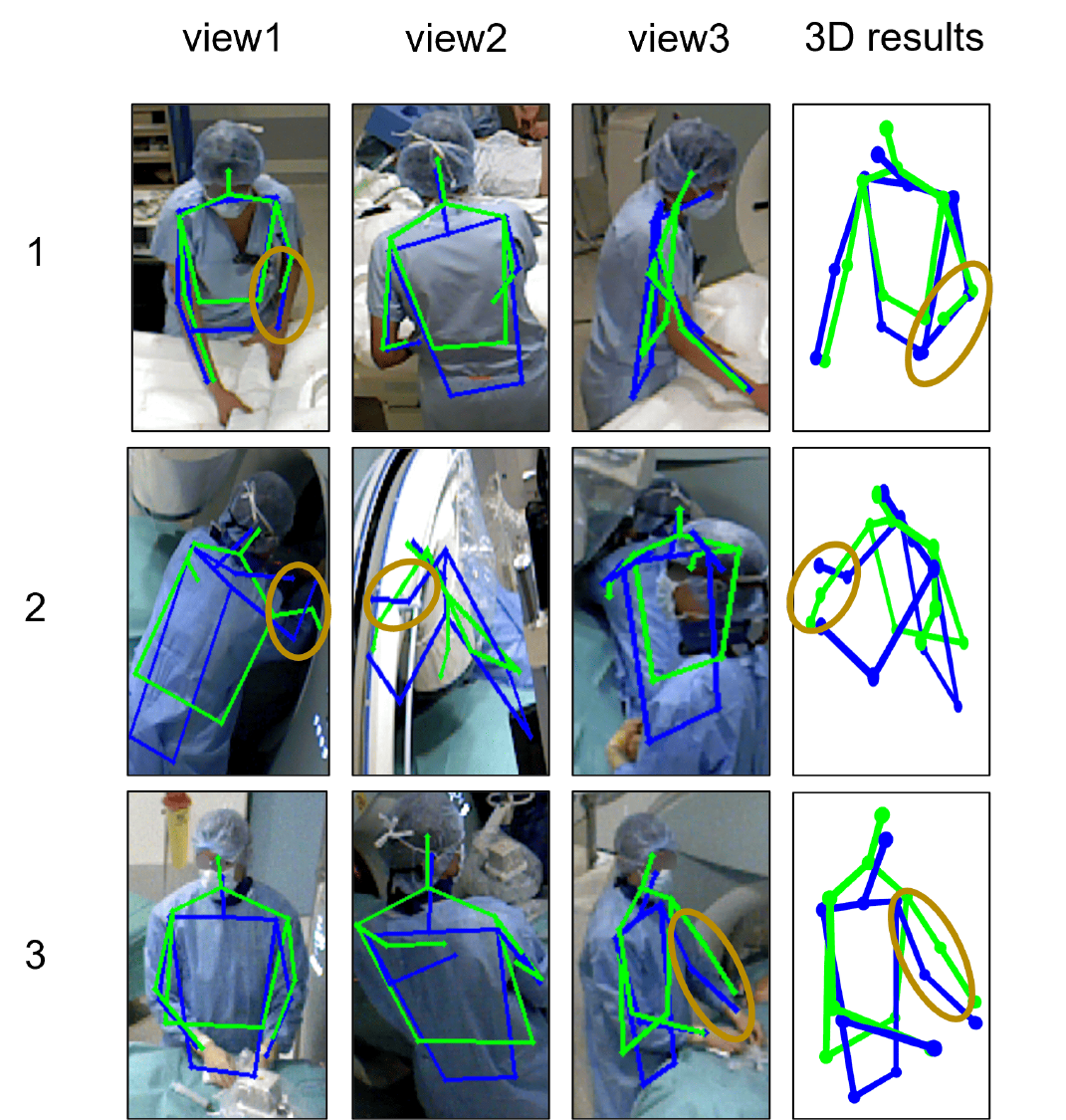}
   \vspace{-0.1cm}
   \caption{Three examples of 3D human pose estimation on MVOR. Blue lines represent predictions, green lines represent ground truth. The first three columns show 2D projections from different views, and the fourth column shows the 3D pose results.}
   \label{fig:mvor}
   \vspace{-0.3cm}
\end{figure}

\begin{figure}[!t]
   \centering
   \includegraphics[width=.9\linewidth]{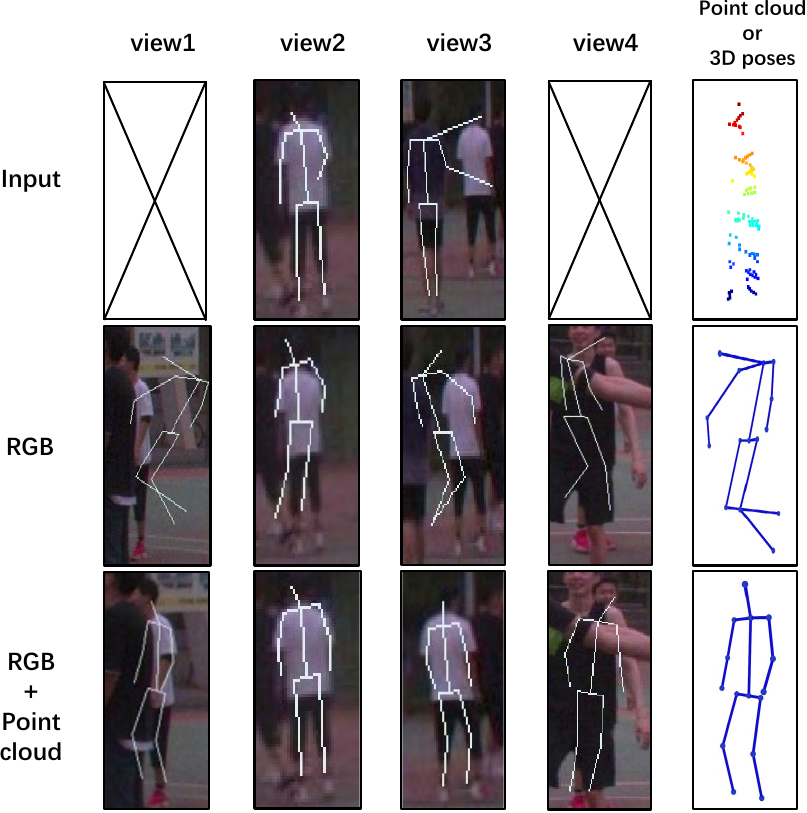}
   \vspace{-0.1cm}
   \caption{Qualitative illustration on the BasketBall dataset from different input modalities. The first row shows 2D pose estimations (missing where not estimable) and point clouds. The second row displays results from using only RGB input, with 2D poses projected from the estimated 3D poses. The third row presents results from using both RGB and point cloud inputs.}
   \label{fig:basketball}
   \vspace{-0.5cm}
\end{figure}

\renewcommand\arraystretch{1}
\begin{table}[t]\footnotesize
   \caption{Comparison of different 3D pose estimation methods on Panoptic and BasketBallSync.}
   \label{tab:pose1}
   \vspace{-0.1cm}
   \centering
   \begin{threeparttable}
      \resizebox{ \linewidth}{!}{
      \begin{tabular}{l l cc cc}
      % \hline
      \toprule
      \multirow{2}{*}{\textbf{Methods}} & \multirow{2}{*}{\textbf{\makecell[l]{Training\\Manner}}} & \multicolumn{2}{c}{\textbf{Panoptic}} & \multicolumn{2}{c}{\textbf{BasketBallSync}} \\ \cmidrule(lr){3-4} \cmidrule(lr){5-6} %\cline{5-6} 
      \multicolumn{1}{c}{}                                  &                                      & MPJPE            & PA-MPJPE            & MPJPE               & PA-MPJPE               \\ \midrule
      MvP \cite{2021direct}                                                   & supervised                                 & 25.78                 & 25.02                    &  291.11                   &  240.32                      \\
      PlanePose \cite{2021multi}                                             & supervised                                  & 18.54                 & 11.92                    &  119.55                   &  65.80                      \\
      VoxelPose(PRN) \cite{2020voxelpose}                              & supervised                                 & 14.91                        &11.88                    &  40.80                   &  34.34                     \\
      LiCamPose                                                   & supervised                                  &  \textbf{14.44}                & \textbf{11.61}                    &  \textbf{31.84}      &  \textbf{27.36}                      \\ \midrule
      DLT \cite{remelli2020lightweight}                               & unsupervised                                  & 38.26                 & 44.77                    & 101.49                 & 62.77                        \\
      LiCamPose                                                      & unsupervised\tnote{$\dagger$}                                  & \textbf{22.00}                 & \textbf{15.96}                    &  \textbf{72.92}                   &  \textbf{62.72}                      \\ \bottomrule
      \end{tabular}}  
      \begin{tablenotes}\scriptsize
         \item[$\dagger$] Pretrained on BasketBallSync during Panoptic evaluating and on Panoptic \\ during BasketBallSync testing. It is the same for Table \ref{tab:pose2} and \ref{tab:ablation}.
      \end{tablenotes}
   \end{threeparttable}
   \vspace{-0.3cm}
\end{table}

\renewcommand\arraystretch{1}
\begin{table}[t]\footnotesize
   \caption{Comparison of Different Input Modalities for LiCamPose on Panoptic and BasketBallSync.}
   \label{tab:pose2}
   \vspace{-0.1cm}
   \resizebox{ \linewidth}{!}{
   \begin{tabular}{l l cc cc}
   \toprule
   \multirow{2}{*}{\textbf{\makecell[l]{Input\\Modality}}} & \multirow{2}{*}{\textbf{\makecell[l]{Training\\Manenr}}} & \multicolumn{2}{c}{\textbf{Panoptic}} & \multicolumn{2}{c}{\textbf{BasketBallSync}} \\ \cmidrule(lr){3-4} \cmidrule(lr){5-6} %\cline{3-6} 
   \multicolumn{1}{c}{}                                                                          &                              & MPJPE        & PA-MPJPE       & MPJPE          & PA-MPJPE          \\ \midrule
   point cloud   & supervised                         & 164.30             & 148.78           &  136.01              & 123.22                  \\
   RGB & supervised                        & 14.91             & 11.88               &  40.80         & 34.34                  \\
   RGB + point cloud & supervised                         & \textbf{14.44}             & \textbf{11.61}               &  \textbf{31.84}              & \textbf{27.36}                  \\ \midrule
   RGB & unsupervised                & 25.72             & 17.24               &  263.05              & 183.85                 \\
   RGB + point cloud & unsupervised                      & \textbf{22.00}             & \textbf{15.96}               &  \textbf{72.92}              & \textbf{62.72}                  \\ \bottomrule
   \end{tabular}}
   \vspace{-0.5cm}
\end{table}

\begin{table*}[!t]
   \centering
   \caption{Comparison of different scene setups. We simulated four different scene setups using the SyncHuman generator. It shows the performance of LiCamPose when pre-trained on different scenes and directly applied to Panoptic Studio.} \label{tab:sync}
   \vspace{-0.1cm}
   \begin{threeparttable}
      \begin{tabular}{l c llc c}
      \toprule
      \multirow{2}{*}{\textbf{Scene's Size}\tnote{1}} & \multicolumn{3}{c}{\textbf{Sensors' Setting}\tnote{2}}                                    & \multicolumn{2}{c}{\textbf{Metric}} \\ \cmidrule(lr){2-4} \cmidrule(lr){5-6}
                                    & Number\tnote{3} & \multicolumn{1}{l}{Position} & \multicolumn{1}{l}{Orientation} & MPJPE      & PA-MPJPE      \\ \midrule
      BaseketBall                               &  4     &  BaseketBall                            &  BaseketBall                                & 82.61           & 93.41              \\
      BaseketBall                               &  5     &  others                          &  Panoptic                                & 77.90           & 60.42              \\
      Panoptic                               &  5     &  others                          &  Panoptic                                & 26.45           & 22.56              \\
      Panoptic                               &  5     &  Panoptic                            &  Panoptic                                & \textbf{23.15}  & \textbf{17.63}              \\ \bottomrule
      \end{tabular}
      \begin{tablenotes}\footnotesize 
         \item[1] Scene sizes are denoted with the names of the corresponding datasets.
         \item[2] Sensor settings are denoted with the names of the corresponding datasets. "Others" refers to configurations with different settings.
         \item[3] BasketBall has 4 groups of sensors, and Panoptic Studio has 5 groups of sensors. 
      \end{tablenotes}
   \end{threeparttable}
   \vspace{-0.3cm}
\end{table*}

\begin{table}[!t]
   \small
   \centering
   \caption{Comparison of different unsupervised training strategies.} \label{tab:ablation}
   \vspace{-0.1cm}
   \resizebox{\linewidth}{!}{
   \begin{tabular}{c cc cc cc}
   \toprule
    \multicolumn{2}{c}{\textbf{Unsupervised loss}} & \multicolumn{2}{c}{\textbf{Panoptic}} & \multicolumn{2}{c}{\textbf{BasketBallSync}} \\ \cmidrule(lr){1-2} \cmidrule(lr){3-4} \cmidrule(lr){5-6} %\cline{2-7} 
    $\mathcal{L}_{\text{3D}}$          & $\mathcal{L}_{\text{prior}}$         & MPJPE        & PA-MPJPE       & MPJPE          & PA-MPJPE          \\ \midrule
    \usym{2717}      & \usym{2717}              & 29.14             & 16.48               & 84.49               & 75.55                  \\
    \usym{2713}      & \usym{2717}              & 28.26             & 25.85               & 77.38               & 68.84                  \\
    \usym{2717}      & \usym{2713}               & 22.88             & 24.69               & 76.45               & 65.54                  \\
    \usym{2713}      & \usym{2713}              & \textbf{22.00}             & \textbf{15.96}               &  \textbf{72.92}              & \textbf{62.72}                  \\ \bottomrule
   \end{tabular}}
   \vspace{-0.5cm}
\end{table}

\subsection{3D Pose Estimation Analysis} \label{pose}
% In this section, we evaluate the 3D human pose estimation results of different approaches. For PlaneSweepPose \cite{2021multi}, VoxelPose \cite{2020voxelpose} and LiCamPose in supervised manner, we utilize the groundtruth location of 
% each person into the process of training and testing. As to MvP \cite{2021direct}, it is difficult to add the groundtruth detection information into the network due to its architecture design, which also means that it cannot utilize the depth or point cloud 
% informaion directly. For fairness, we just evaluate the MPJPE@500 metrics (only considering per joint error less than 500 mm) for the matched person as to MvP. 
In this section, we evaluate the 3D human pose estimation results using different approaches. For PlaneSweepPose \cite{2021multi}, VoxelPose \cite{2020voxelpose}, and supervised LiCamPose, we incorporate ground truth person locations during both training and testing phases. However, for MvP \cite{2021direct}, its architecture does not accommodate direct integration of ground truth detection information, making it unable to utilize depth or point cloud data directly. For fairness, we evaluate only the MPJPE@500 metric (considering only per-joint errors less than 500 mm) for the matched person in the case of MvP.

% We evaluate these approaches on Panoptic Studio and BasketballSync as small scene and large scene respectively. Table \ref{tab:pose1} shows that our approach outperforms them in terms of MPJPE and PA-MPJPE especially in large 
% scene's setting. It is worth noting that our approach with unsupervised learning manner can even outperform supervised RGB inputs multi-view approaches.
We evaluate these approaches on Panoptic Studio and BasketBallSync datasets, representing small and large scenes respectively. Table \ref{tab:pose1} demonstrates that LiCamPose surpasses them in terms of both MPJPE and PA-MPJPE, particularly in the large scene setting. Notably, unsupervised learning LiCamPose even outperforms some supervised multi-view approaches with RGB inputs.

% We assess our pipeline on MVOR, and our method in unsupervised manner achieved great performance which even outperforms groundtruth in some frames (Figure \ref{fig:mvor}).
We evaluate LiCamPose using unsupervised domain adaption on MVOR, and it achieves impressive performance, occasionally outperforming ground truth in some frames (Figure \ref{fig:mvor}).

% Additionally, we analyze the performance with different modals inputs (point cloud, RGB, and both of them) in volumetric architecture. 
% Table \ref{tab:pose2} illustrates that the Livox point cloud information alone
% is unable to accurately extract 3D human skeleton due to its sparsity, and cannot be adopted in unsupervised manner without 2D pseudo labels from heatmaps. 
% However, when combined with RGB information, the performance is greatly enhanced.
% Figure \ref{fig:basketball} demonstrates that multi-modal inputs increase the model's tolerance to 2D pose estimation error, including severe situations like miss-prediction or predicting on a wrong person, 
% during unsupervised domain adaption training. 
Additionally, we analyze the performance using different modal inputs (point cloud, RGB, and both) in volumetric representation. Table \ref{tab:pose2} illustrates that using Livox point cloud information alone struggles to accurately extract the 3D human skeleton due to its sparsity, making it unsuitable for unsupervised learning without 2D pseudo-labels from heatmaps. However, combining it with RGB information significantly enhances performance. Figure \ref{fig:basketball} demonstrates that multi-modal inputs improve the model's robustness to 2D pose estimation errors, even in challenging scenarios such as mispredictions or incorrect person predictions, during unsupervised domain adaptation training.

\subsection{Unsupervised Domain Adaption}
In this section, we conduct a detailed analysis of synthetic dataset settings and ablation studies on the unsupervised training Losses.

\subsubsection{Synthetic Setting}
Tu et al. \cite{2020voxelpose} conducted an experiment training VoxelPose with a dataset's camera configuration and human poses from other sources, then evaluating the model directly on another dataset. In this section, we further analyze the impact of scene settings in volumetric architecture. We simulated four different scene setups using the SyncHuman generator, varying sensor positions and orientations to modify camera parameters and scene sizes. Table \ref{tab:sync} demonstrates that the closer the scene setups resemble each other, the better the model performs. This suggests that in practical applications, aligning synthetic data with the sensor arrangement and scene range of the target dataset can enhance the performance of unsupervised domain adaptation or direct inference in new scenarios.

\subsubsection{Ablation Study on Unsupervised Training Losses} 
For unsupervised domain adaptation, pseudo 2D pose supervision serves as the baseline necessity. Additionally, we incorporate an interpretable human prior loss and a pseudo 3D pose loss (selected based on entropy values) to aid in learning. Thus, we conduct an ablation study to analyze the impact of these two losses. According to Table \ref{tab:ablation}, the entropy-selected pseudo 3D pose loss improves performance, and the prior loss ensures predicted poses remain within a reasonable action range. Therefore, we adopt both losses as part of our effective training strategy. 
% Moreover, these losses enhance robustness to 2D pose estimation errors (Figure \ref{fig:basketball2}).

\begin{figure}[!t]
   \centering
   \includegraphics[width=0.85\linewidth]{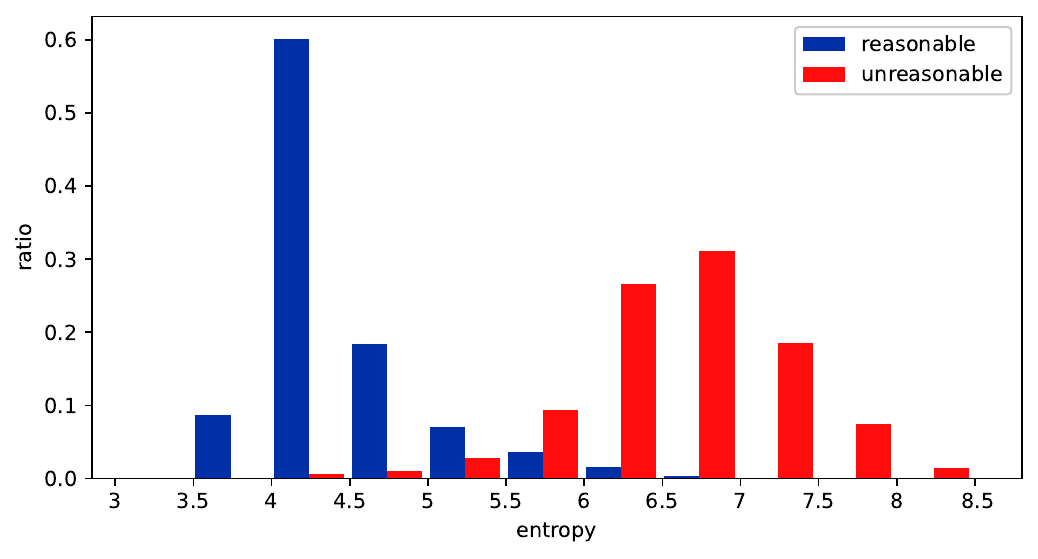}
   \vspace{-0.1cm}
   \caption{Entropy distributions of reasonable and unreasonable predicted 3D poses.}
   \label{fig:entropy}
   \vspace{-0.5cm}
\end{figure}

\subsection{Entropy Analysis} \label{entropy_ana}
To explore the relationship between entropy values and pose rationality, we categorized some 3D poses predicted by LiCamPose as either reasonable or unreasonable manually. Figure \ref{fig:entropy} illustrates the distribution of entropy values for these categories. Natural division of poses occurs based on an entropy threshold. Therefore, the average entropy value can serve as an indicator of a model's performance on an unlabeled dataset. 
% It's important to note that reducing entropy values post unsupervised learning is challenging without adequate pretraining. Hence, entropy-based assessment is particularly relevant before unsupervised adaptation.

% \section{Limitation}
% The LiCamPose pipeline performs well in both supervised training or unsupervised domain adaption. However, it still has some problems to be addressed.
% 1) The method does not incorporate temporal information. 2) The unsupervised adaption performance on single view dataset is suboptimal. 3) The synthetic dataset currently lacks complex backgrounds and other interpretable objects.

\section{Conclusion}
%%%%%%%%% BODY TEXT
%  In this paper, we propose a pipeline called LiCamPose for 3D human pose estimation using multi-view RGB and pointcloud information.
%  To train the model in unsupervised manner, we develop a synthetic generator SyncHuman and design an efficient unsupervised loss for domain adaption learning.
%  We conduct extensive experiments on both synthetic and real-world datasets. We analyze the powerful learning ability of this pipeline for 3D human pose estimation. 
%  Besides, we study the impact of scene settings and unsupervised training losses. Furthermore, we investigate the inherent relationship between entropy value and pose plausibility. 
%  We believe that our approach can be a good baseline for large scene's multi-view LiDAR-RGB 3D human pose estimation.
In this paper, we propose LiCamPose, a pipeline for 3D human pose estimation using multi-view RGB and sparse point cloud data. To facilitate unsupervised training, we introduce SyncHuman, a synthetic data generator, and develop specific unsupervised training losses for domain adaptation. Through extensive experiments on synthetic and real-world datasets, we demonstrate the robust learning capabilities of our pipeline for 3D human pose estimation. Additionally, we investigate the influence of scene settings and unsupervised training losses, and explore the relationship between entropy values and pose plausibility. In future work, we aim to expand SyncHuman's scenarios and integrate temporal information to further enhance estimation accuracy.

%%%%%%%%% REFERENCES
{\small
\bibliographystyle{ieee_fullname}
\bibliography{main}

@String(ICPR = {Int. Conf. Pattern Recog.})

@String(AAAI = {AAAI})

@String(ICPR  = {ICPR})

@inproceedings{2019learnable,
  title={Learnable triangulation of human pose},
  author={Iskakov, Karim and Burkov, Egor and Lempitsky, Victor and Malkov, Yury},
  booktitle={Proceedings of the IEEE/CVF International Conference on Computer Vision},
  pages={7718--7727},
  year={2019}
}

@inproceedings{2019fast,
  title={Fast and robust multi-person 3{D} pose estimation from multiple views},
  author={Dong, Junting and Jiang, Wen and Huang, Qixing and Bao, Hujun and Zhou, Xiaowei},
  booktitle={Proceedings of the IEEE/CVF Conference on Computer Vision and Pattern Recognition},
  pages={7792--7801},
  year={2019}
}

@article{bekhtaoui2020view,
  title={View invariant human body detection and pose estimation from multiple depth sensors},
  author={Bekhtaoui, Walid and Sa, Ruhan and Teixeira, Brian and Singh, Vivek and Kirchberg, Klaus and Chang, Yao-jen and Kapoor, Ankur},
  journal={arXiv preprint arXiv:2005.04258},
  year={2020}
}

@inproceedings{2020voxelpose,
  title={{VoxelPose}: Towards multi-camera 3{D} human pose estimation in wild environment},
  author={Tu, Hanyue and Wang, Chunyu and Zeng, Wenjun},
  booktitle={Proceedings of the European Conference on Computer Vision},
  pages={197--212},
  year={2020}
}

@inproceedings{2021multi,
  title={Multi-view multi-person 3{D} pose estimation with {Plane Sweep Stereo}},
  author={Lin, Jiahao and Lee, Gim Hee},
  booktitle={Proceedings of the IEEE/CVF Conference on Computer Vision and Pattern Recognition},
  pages={11886--11895},
  year={2021}
}

@inproceedings{2021graph,
  title={Graph-based 3{D} multi-person pose estimation using multi-view images},
  author={Wu, Size and Jin, Sheng and Liu, Wentao and Bai, Lei and Qian, Chen and Liu, Dong and Ouyang, Wanli},
  booktitle={Proceedings of the IEEE/CVF International Conference on Computer Vision},
  pages={11148--11157},
  year={2021}
}

@article{2021direct,
  title={Direct multi-view multi-person 3{D} pose estimation},
  author={Zhang, Jianfeng and Cai, Yujun and Yan, Shuicheng and Feng, Jiashi and others},
  journal={Advances in Neural Information Processing Systems},
  volume={34},
  pages={13153--13164},
  year={2021}
}

@inproceedings{2021tessetrack,
  title={{T}essetrack: End-to-end learnable multi-person articulated 3{D} pose tracking},
  author={Reddy, N Dinesh and Guigues, Laurent and Pishchulin, Leonid and Eledath, Jayan and Narasimhan, Srinivasa G},
  booktitle={Proceedings of the IEEE/CVF Conference on Computer Vision and Pattern Recognition},
  pages={15190--15200},
  year={2021}
}

@inproceedings{2022faster,
  title={{F}aster {V}oxel{P}ose: Real-time 3D Human Pose Estimation by Orthographic Projection},
  author={Ye, Hang and Zhu, Wentao and Wang, Chunyu and Wu, Rujie and Wang, Yizhou},
  booktitle={Proceedings of the European Conference on Computer Vision},
  pages={142--159},
  year={2022}
}

@inproceedings{lidarcap,
  title={{Lidarcap}: Long-range marker-less 3d human motion capture with lidar point clouds},
  author={Li, Jialian and Zhang, Jingyi and Wang, Zhiyong and Shen, Siqi and Wen, Chenglu and Ma, Yuexin and Xu, Lan and Yu, Jingyi and Wang, Cheng},
  booktitle={Proceedings of the IEEE/CVF Conference on Computer Vision and Pattern Recognition},
  pages={20502--20512},
  year={2022}
}

@article{vitpose,
  title={{Vitpose}: Simple vision transformer baselines for human pose estimation},
  author={Xu, Yufei and Zhang, Jing and Zhang, Qiming and Tao, Dacheng},
  journal={arXiv preprint arXiv:2204.12484},
  year={2022}
}

@article{alphapose,
  title={{Alphapose}: Whole-body regional multi-person pose estimation and tracking in real-time},
  author={Fang, Hao-Shu and Li, Jiefeng and Tang, Hongyang and Xu, Chao and Zhu, Haoyi and Xiu, Yuliang and Li, Yong-Lu and Lu, Cewu},
  journal={IEEE Transactions on Pattern Analysis and Machine Intelligence},
  year={2022}
}

@inproceedings{openpose,
  title={Realtime multi-person {2D} pose estimation using part affinity fields},
  author={Cao, Zhe and Simon, Tomas and Wei, Shih-En and Sheikh, Yaser},
  booktitle={Proceedings of the IEEE Conference on Computer Vision and Pattern Recognition},
  pages={7291--7299},
  year={2017}
}

@inproceedings{hrnet,
  title={Deep high-resolution representation learning for human pose estimation},
  author={Sun, Ke and Xiao, Bin and Liu, Dong and Wang, Jingdong},
  booktitle={Proceedings of the IEEE/CVF Conference on Computer Vision and Pattern Recognition},
  pages={5693--5703},
  year={2019}
}

@inproceedings{mpii,
  title={{2D} human pose estimation: New benchmark and state of the art analysis},
  author={Andriluka, Mykhaylo and Pishchulin, Leonid and Gehler, Peter and Schiele, Bernt},
  booktitle={Proceedings of the IEEE Conference on Computer Vision and Pattern Recognition},
  pages={3686--3693},
  year={2014}
}

@inproceedings{pose2seg,
  title={{Pose2seg}: Detection free human instance segmentation},
  author={Zhang, Song-Hai and Li, Ruilong and Dong, Xin and Rosin, Paul and Cai, Zixi and Han, Xi and Yang, Dingcheng and Huang, Haozhi and Hu, Shi-Min},
  booktitle={Proceedings of the IEEE/CVF Conference on Computer Vision and Pattern Recognition},
  pages={889--898},
  year={2019}
}

@inproceedings{COCO,
  title={Whole-body human pose estimation in the wild},
  author={Jin, Sheng and Xu, Lumin and Xu, Jin and Wang, Can and Liu, Wentao and Qian, Chen and Ouyang, Wanli and Luo, Ping},
  booktitle={Proceedings of the European Conference on Computer Vision},
  pages={196--214},
  year={2020}
}

@article{human36m,
  title={{Human3.6M}: Large scale datasets and predictive methods for {3D} human sensing in natural environments},
  author={Ionescu, Catalin and Papava, Dragos and Olaru, Vlad and Sminchisescu, Cristian},
  journal={IEEE Transactions on Pattern Analysis and Machine Intelligence},
  volume={36},
  number={7},
  pages={1325--1339},
  year={2013}
}

@inproceedings{panoptic,
  title={{Panoptic} studio: A massively multiview system for social motion capture},
  author={Joo, Hanbyul and Liu, Hao and Tan, Lei and Gui, Lin and Nabbe, Bart and Matthews, Iain and Kanade, Takeo and Nobuhara, Shohei and Sheikh, Yaser},
  booktitle={Proceedings of the IEEE International Conference on Computer Vision},
  pages={3334--3342},
  year={2015}
}

@inproceedings{3dps,
  title={{3D} pictorial structures for multiple human pose estimation},
  author={Belagiannis, Vasileios and Amin, Sikandar and Andriluka, Mykhaylo and Schiele, Bernt and Navab, Nassir and Ilic, Slobodan},
  booktitle={Proceedings of the IEEE Conference on Computer Vision and Pattern Recognition},
  pages={1669--1676},
  year={2014}
}

@inproceedings{3dpw,
  title={Recovering accurate {3D} human pose in the wild using imus and a moving camera},
  author={Von Marcard, Timo and Henschel, Roberto and Black, Michael J and Rosenhahn, Bodo and Pons-Moll, Gerard},
  booktitle={Proceedings of the European Conference on Computer Vision},
  pages={601--617},
  year={2018}
}

@article{mvor,
  title={MVOR: A multi-view RGB-D operating room dataset for {2D} and {3D} human pose estimation},
  author={Srivastav, Vinkle and Issenhuth, Thibaut and Kadkhodamohammadi, Abdolrahim and de Mathelin, Michel and Gangi, Afshin and Padoy, Nicolas},
  journal={arXiv preprint arXiv:1808.08180},
  year={2018}
}

@article{bevfusion,
  title={{BEVFusion}: Multi-Task Multi-Sensor Fusion with Unified Bird's-Eye View Representation},
  author={Liu, Zhijian and Tang, Haotian and Amini, Alexander and Yang, Xinyu and Mao, Huizi and Rus, Daniela and Han, Song},
  journal={arXiv preprint arXiv:2205.13542},
  year={2022}
}

@inproceedings{pointpillars,
  title={{PointPillars}: Fast encoders for object detection from point clouds},
  author={Lang, Alex H and Vora, Sourabh and Caesar, Holger and Zhou, Lubing and Yang, Jiong and Beijbom, Oscar},
  booktitle={Proceedings of the IEEE/CVF Conference on Computer Vision and Pattern Recognition},
  pages={12697--12705},
  year={2019}
}

@inproceedings{votenet,
  title={Deep hough voting for {3D} object detection in point clouds},
  author={Qi, Charles R and Litany, Or and He, Kaiming and Guibas, Leonidas J},
  booktitle={Proceedings of the IEEE/CVF International Conference on Computer Vision},
  pages={9277--9286},
  year={2019}
}

@inproceedings{amass,
  title={{AMASS}: Archive of motion capture as surface shapes},
  author={Mahmood, Naureen and Ghorbani, Nima and Troje, Nikolaus F and Pons-Moll, Gerard and Black, Michael J},
  booktitle={Proceedings of the IEEE/CVF International Conference on Computer Vision},
  pages={5442--5451},
  year={2019}
}

@inproceedings{bigalke2022domain,
  title={Domain adaptation through anatomical constraints for {3D} human pose estimation under the cover},
  author={Bigalke, Alexander and Hansen, Lasse and Diesel, Jasper and Heinrich, Mattias P},
  booktitle={International Conference on Medical Imaging with Deep Learning},
  pages={173--187},
  year={2022}
}

@article{gerats20223d,
  title={{3D} human pose estimation in multi-view operating room videos using differentiable camera projections},
  author={Gerats, Beerend GA and Wolterink, Jelmer M and Broeders, Ivo AMJ},
  journal={Computer Methods in Biomechanics and Biomedical Engineering: Imaging \& Visualization},
  pages={1--9},
  year={2022}
}

@article{hansen2019fusing,
  title={Fusing information from multiple {2D} depth cameras for {3D} human pose estimation in the operating room},
  author={Hansen, Lasse and Siebert, Marlin and Diesel, Jasper and Heinrich, Mattias P},
  journal={International Journal of Computer Assisted Radiology and Surgery},
  volume={14},
  pages={1871--1879},
  year={2019}
}

@inproceedings{surreal,
  title={Learning from synthetic humans},
  author={Varol, Gul and Romero, Javier and Martin, Xavier and Mahmood, Naureen and Black, Michael J and Laptev, Ivan and Schmid, Cordelia},
  booktitle={Proceedings of the IEEE Conference on computer vision and pattern recognition},
  pages={109--117},
  year={2017}
}

@inproceedings{cape,
  title={Learning to dress {3D} people in generative clothing},
  author={Ma, Qianli and Yang, Jinlong and Ranjan, Anurag and Pujades, Sergi and Pons-Moll, Gerard and Tang, Siyu and Black, Michael J},
  booktitle={Proceedings of the IEEE/CVF Conference on Computer Vision and Pattern Recognition},
  pages={6469--6478},
  year={2020}
}

@inproceedings{carla,
  title={{CARLA}: An open urban driving simulator},
  author={Dosovitskiy, Alexey and Ros, German and Codevilla, Felipe and Lopez, Antonio and Koltun, Vladlen},
  booktitle={Conference on Robot Learning},
  pages={1--16},
  year={2017}
}

@article{peoplesanspeople,
  title={{PeopleSansPeople}: a synthetic data generator for human-centric computer vision},
  author={Ebadi, Salehe Erfanian and Jhang, You-Cyuan and Zook, Alex and Dhakad, Saurav and Crespi, Adam and Parisi, Pete and Borkman, Steven and Hogins, Jonathan and Ganguly, Sujoy},
  journal={arXiv preprint arXiv:2112.09290},
  year={2021}
}

@inproceedings{2022uncertainty,
  title={Uncertainty-aware adaptation for self-supervised {3D} human pose estimation},
  author={Kundu, Jogendra Nath and Seth, Siddharth and YM, Pradyumna and Jampani, Varun and Chakraborty, Anirban and Babu, R Venkatesh},
  booktitle={Proceedings of the IEEE/CVF Conference on Computer Vision and Pattern Recognition},
  pages={20448--20459},
  year={2022}
}

@inproceedings{2019self,
  title={Self-supervised learning of {3D} human pose using multi-view geometry},
  author={Kocabas, Muhammed and Karagoz, Salih and Akbas, Emre},
  booktitle={Proceedings of the IEEE/CVF Conference on Computer Vision and Pattern Recognition},
  pages={1077--1086},
  year={2019}
}

@inproceedings{srivastav2020self,
  title={Self-supervision on unlabelled or data for multi-person {2D/3D} human pose estimation},
  author={Srivastav, Vinkle and Gangi, Afshin and Padoy, Nicolas},
  booktitle={Medical Image Computing and Computer Assisted Intervention},
  pages={761--771},
  year={2020}
}

@inproceedings{kundu2020unsupervised,
  title={Unsupervised cross-modal alignment for multi-person {3D} pose estimation},
  author={Kundu, Jogendra Nath and Revanur, Ambareesh and Waghmare, Govind Vitthal and Venkatesh, Rahul Mysore and Babu, R Venkatesh},
  booktitle={Proceedings of the European Conference on Computer Vision},
  pages={35--52},
  year={2020}
}

@article{doersch2019sim2real,
  title={{Sim2real} transfer learning for {3D} human pose estimation: motion to the rescue},
  author={Doersch, Carl and Zisserman, Andrew},
  journal={Advances in Neural Information Processing Systems},
  volume={32},
  year={2019}
}

@inproceedings{zhang2019unsupervised,
  title={Unsupervised domain adaptation for {3D} human pose estimation},
  author={Zhang, Xiheng and Wong, Yongkang and Kankanhalli, Mohan S and Geng, Weidong},
  booktitle={Proceedings of the 27th ACM International Conference on Multimedia},
  pages={926--934},
  year={2019}
}

@inproceedings{garau2021deca,
  title={{DECA}: Deep viewpoint-Equivariant human pose estimation using Capsule Autoencoders},
  author={Garau, Nicola and Bisagno, Niccolo and Br{\'o}dka, Piotr and Conci, Nicola},
  booktitle={Proceedings of the IEEE/CVF International Conference on Computer Vision},
  pages={11677--11686},
  year={2021}
}

@inproceedings{haque2016towards,
  title={Towards viewpoint invariant {3D} human pose estimation},
  author={Haque, Albert and Peng, Boya and Luo, Zelun and Alahi, Alexandre and Yeung, Serena and Fei-Fei, Li},
  booktitle={Proceedings of the European Conference on Computer Vision},
  pages={160--177},
  year={2016}
}

@article{guo2017towards,
  title={Towards good practices for deep {3D} hand pose estimation},
  author={Guo, Hengkai and Wang, Guijin and Chen, Xinghao and Zhang, Cairong},
  journal={arXiv preprint arXiv:1707.07248},
  year={2017}
}

@inproceedings{zhang2021sequential,
  title={Sequential {3D} Human Pose Estimation Using Adaptive Point Cloud Sampling Strategy.},
  author={Zhang, Zihao and Hu, Lei and Deng, Xiaoming and Xia, Shihong},
  booktitle={International Joint Conferences on Artificial Intelligence Organization},
  pages={1330--1337},
  year={2021}
}

@inproceedings{moon2018v2v,
  title={{V2V-Posenet}: Voxel-to-voxel prediction network for accurate {3D} hand and human pose estimation from a single depth map},
  author={Moon, Gyeongsik and Chang, Ju Yong and Lee, Kyoung Mu},
  booktitle={Proceedings of the IEEE conference on Computer Vision and Pattern Recognition},
  pages={5079--5088},
  year={2018}
}

@inproceedings{bashirov2021real,
  title={Real-time rgbd-based extended body pose estimation},
  author={Bashirov, Renat and Ianina, Anastasia and Iskakov, Karim and Kononenko, Yevgeniy and Strizhkova, Valeriya and Lempitsky, Victor and Vakhitov, Alexander},
  booktitle={Proceedings of the IEEE/CVF Winter Conference on Applications of Computer Vision},
  pages={2807--2816},
  year={2021}
}

@inproceedings{zheng2022multi,
  title={Multi-modal {3D} human pose estimation with {2D} weak supervision in autonomous driving},
  author={Zheng, Jingxiao and Shi, Xinwei and Gorban, Alexander and Mao, Junhua and Song, Yang and Qi, Charles R and Liu, Ting and Chari, Visesh and Cornman, Andre and Zhou, Yin and others},
  booktitle={Proceedings of the IEEE/CVF Conference on Computer Vision and Pattern Recognition},
  pages={4478--4487},
  year={2022}
}

@inproceedings{martinez2020residual,
  title={Residual pose: A decoupled approach for depth-based {3D} human pose estimation},
  author={Mart{\'\i}nez-Gonz{\'a}lez, Angel and Villamizar, Michael and Can{\'e}vet, Olivier and Odobez, Jean-Marc},
  booktitle={IEEE/RSJ International Conference on Intelligent Robots and Systems},
  pages={10313--10318},
  year={2020}
}

@inproceedings{piergiovanni20214d,
  title={{4D-net} for learned multi-modal alignment},
  author={Piergiovanni, AJ and Casser, Vincent and Ryoo, Michael S and Angelova, Anelia},
  booktitle={Proceedings of the IEEE/CVF International Conference on Computer Vision},
  pages={15435--15445},
  year={2021}
}

@inproceedings{liang2018deep,
  title={Deep continuous fusion for multi-sensor {3D} object detection},
  author={Liang, Ming and Yang, Bin and Wang, Shenlong and Urtasun, Raquel},
  booktitle={Proceedings of the European Conference on Computer Vision},
  pages={641--656},
  year={2018}
}

@inproceedings{mvdet,
  title={Multiview detection with feature perspective transformation},
  author={Hou, Yunzhong and Zheng, Liang and Gould, Stephen},
  booktitle={Proceedings of the European Conference on Computer Vision},
  pages={1--18},
  year={2020}
}

@article{gholami2022self,
  title={Self-supervised {3D} human pose estimation from video},
  author={Gholami, Mohsen and Rezaei, Ahmad and Rhodin, Helge and Ward, Rabab and Wang, Z Jane},
  journal={Neurocomputing},
  volume={488},
  pages={97--106},
  year={2022}
}

@inproceedings{lin2014microsoft,
  title={{Microsoft COCO}: Common objects in context},
  author={Lin, Tsung-Yi and Maire, Michael and Belongie, Serge and Hays, James and Perona, Pietro and Ramanan, Deva and Doll{\'a}r, Piotr and Zitnick, C Lawrence},
  booktitle={Proceedings of the European Conference on Computer Vision},
  pages={740--755},
  year={2014}
}

@article{loper2015smpl,
  title={{SMPL}: A skinned multi-person linear model},
  author={Loper, Matthew and Mahmood, Naureen and Romero, Javier and Pons-Moll, Gerard and Black, Michael J},
  journal={ACM Transactions on Graphics},
  volume={34},
  number={6},
  pages={1--16},
  year={2015}
}

@article{kingma2014adam,
  title={{Adam}: A method for stochastic optimization},
  author={Kingma, Diederik P and Ba, Jimmy},
  journal={arXiv preprint arXiv:1412.6980},
  year={2014}
}

@inproceedings{geiger2012we,
  title={Are we ready for autonomous driving? the {KITTI} vision benchmark suite},
  author={Geiger, Andreas and Lenz, Philip and Urtasun, Raquel},
  booktitle={Proceedings of the IEEE Conference on Computer Vision and Pattern Recognition},
  pages={3354--3361},
  year={2012}
}

@article{gower1975generalized,
  title={Generalized procrustes analysis},
  author={Gower, John C},
  journal={Psychometrika},
  volume={40},
  pages={33--51},
  year={1975}
}

@inproceedings{remelli2020lightweight,
  title={Lightweight multi-view {3D} pose estimation through camera-disentangled representation},
  author={Remelli, Edoardo and Han, Shangchen and Honari, Sina and Fua, Pascal and Wang, Robert},
  booktitle={Proceedings of the IEEE/CVF Conference on Computer Vision and Pattern Recognition},
  pages={6040--6049},
  year={2020}
}

@inproceedings{yang2023bevformer,
  title={Bevformer v2: Adapting modern image backbones to bird's-eye-view recognition via perspective supervision},
  author={Yang, Chenyu and Chen, Yuntao and Tian, Hao and Tao, Chenxin and Zhu, Xizhou and Zhang, Zhaoxiang and Huang, Gao and Li, Hongyang and Qiao, Yu and Lu, Lewei and others},
  booktitle={Proceedings of the IEEE/CVF Conference on Computer Vision and Pattern Recognition},
  pages={17830--17839},
  year={2023}
}

@inproceedings{zhang2022flexible,
  title={A Flexible Multi-view Multi-modal Imaging System for Outdoor Scenes},
  author={Zhang, Meng and Guo, Wenxuan and Fan, Bohao and Chen, Yifan and Feng, Jianjiang and Zhou, Jie},
  booktitle={2022 International Conference on 3D Vision (3DV)},
  pages={322--331},
  year={2022},
  organization={IEEE}
}

@inproceedings{AGORA,
  title={{AGORA}: Avatars in geography optimized for regression analysis},
  author={Patel, Priyanka and Huang, Chun-Hao P and Tesch, Joachim and Hoffmann, David T and Tripathi, Shashank and Black, Michael J},
  booktitle={Proceedings of the IEEE/CVF Conference on Computer Vision and Pattern Recognition},
  pages={13468--13478},
  year={2021}
}

@inproceedings{BEDLAM,
  title={{BEDLAM}: A synthetic dataset of bodies exhibiting detailed lifelike animated motion},
  author={Black, Michael J and Patel, Priyanka and Tesch, Joachim and Yang, Jinlong},
  booktitle={Proceedings of the IEEE/CVF Conference on Computer Vision and Pattern Recognition},
  pages={8726--8737},
  year={2023}
}

@inproceedings{BlendMimic3D,
  title={3D Human Pose Estimation with Occlusions: Introducing BlendMimic3D Dataset and GCN Refinement},
  author={Lino, Filipa and Santiago, Carlos and Marques, Manuel},
  booktitle={Proceedings of the IEEE/CVF Conference on Computer Vision and Pattern Recognition},
  pages={4646--4656},
  year={2024}
}

@inproceedings{HPERL,
  title={HPERL: 3d human pose estimation from RGB and lidar},
  author={F{\"u}rst, Michael and Gupta, Shriya TP and Schuster, Ren{\'e} and Wasenm{\"u}ller, Oliver and Stricker, Didier},
  booktitle={2020 25th International Conference on Pattern Recognition (ICPR)},
  pages={7321--7327},
  year={2021},
  organization={IEEE}
}

@inproceedings{cong2023weakly,
  title={Weakly supervised 3d multi-person pose estimation for large-scale scenes based on monocular camera and single lidar},
  author={Cong, Peishan and Xu, Yiteng and Ren, Yiming and Zhang, Juze and Xu, Lan and Wang, Jingya and Yu, Jingyi and Ma, Yuexin},
  booktitle={Proceedings of the AAAI Conference on Artificial Intelligence},
  volume={37},
  number={1},
  pages={461--469},
  year={2023}
}

@inproceedings{zhang2024neighborhood,
  title={Neighborhood-Enhanced 3D Human Pose Estimation with Monocular LiDAR in Long-Range Outdoor Scenes},
  author={Zhang, Jingyi and Mao, Qihong and Hu, Guosheng and Shen, Siqi and Wang, Cheng},
  booktitle={Proceedings of the AAAI Conference on Artificial Intelligence},
  volume={38},
  number={7},
  pages={7169--7177},
  year={2024}
}
}

\clearpage
\setcounter{page}{1}
\maketitlesupplementary

%%%%%%%%% BODY TEXT
\section{Different Scanning Patterns of Point Cloud}
\begin{figure}[!t]
   \centering
   \includegraphics[width=0.9\linewidth]{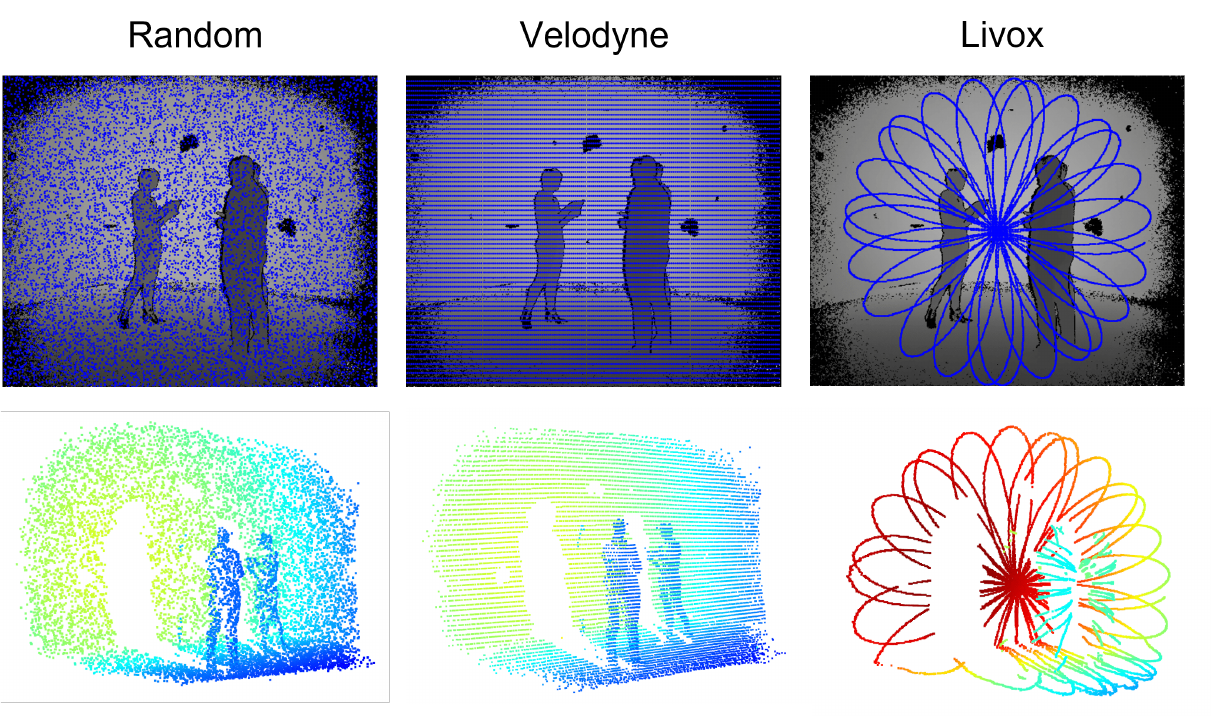}
   \caption{Different scanning patterns of point clouds. All samples shown in this figure are from the same scene, captured at the same time, and contain the same number of points.}
   \label{fig:scanning pattern}
\end{figure}

   % There are different ways to get or scan the point cloud: 1) randomly sampling the depth map; 2) sampling the depth map with multiple 
   % equidistant horizontal lines to imitate Velodyne LiDARs; 3) sampling the depth map with the ``Rose curve'' sampling equation mentioned 
   % in our paper to imitate Livox LiDARs. Figure \ref{fig:scanning pattern} shows that the ``Rose curve'' sampling equation provides minimal information due to its localized concentrated scan. However, Livox LiDARs are more affordable than Velodyne LiDARs, and they 
   % have been used in many applications including surveillance. Besides, our BaseketBall data is captured by Livox LiDARs. 
   % Therefore, we use the Livox scanning pattern to simulate scanning pointcloud in our experiments. 
   There are various methods to obtain or scan the point cloud: 1) randomly sampling the depth map; 2) sampling the depth map using multiple equidistant horizontal lines to mimic Velodyne LiDARs; and 3) sampling the depth map with the "Rose curve" sampling equation as discussed in our paper to replicate Livox LiDARs. Figure \ref{fig:scanning pattern} illustrates that the "Rose curve" sampling equation yields minimal information due to its localized concentrated scan. However, Livox LiDARs are more cost-effective than Velodyne LiDARs and have been employed in numerous applications, including surveillance. Additionally, our BaseketBall dataset is captured using Livox LiDARs. Therefore, we adopt the Livox scanning pattern to simulate the point cloud scanning in our experiments.

\section{BaseketBall} % a qualitive description of the datasets
   % BasketBall is an outdoor dataset capturing a Basketball match with four sensor nodes (each node includes one Livox LiDAR and one RGB camera) in a convergent acquisition. 
   % The dataset is challenging due to large covering, occlusions, and players' dynamic motions (Figure \ref{fig:dataset}). Currently, we have developed an 
   % annotation tool to annotate the players' 3D bounding boxes and IDs. In the future, we will integrate the 3D human keypoints annotation into the tool with the assist of LiCamPose.  
   BasketBall is an outdoor dataset capturing a basketball match using four sensor nodes, each comprising one Livox LiDAR and one RGB camera, in a convergent acquisition setup. The dataset presents challenges due to its extensive coverage, occlusions, and the dynamic motions of the players (Figure \ref{fig:dataset}). We have developed an annotation tool to label the players' 3D bounding boxes and IDs. In the future, we plan to integrate 3D human keypoint annotation into the tool with the assistance of LiCamPose.
% \section{SyncHuman Generator}
%    We can use our synthetic data generator, SyncHuman, to simulate any arrangement of sensors to observe a scene. As demonstrated in the experiments in our paper, using the same scene setting for both training and testing yields better transfer performance. Figure \ref{fig:dataset} compares the datasets we generated, PanopticSync and BasketBallSync, with the original datasets.

\section{The Detailed Structure of V2V-Net and Fusion-Net}
\begin{figure*}[t]
   \centering
   \includegraphics[width=0.9\linewidth]{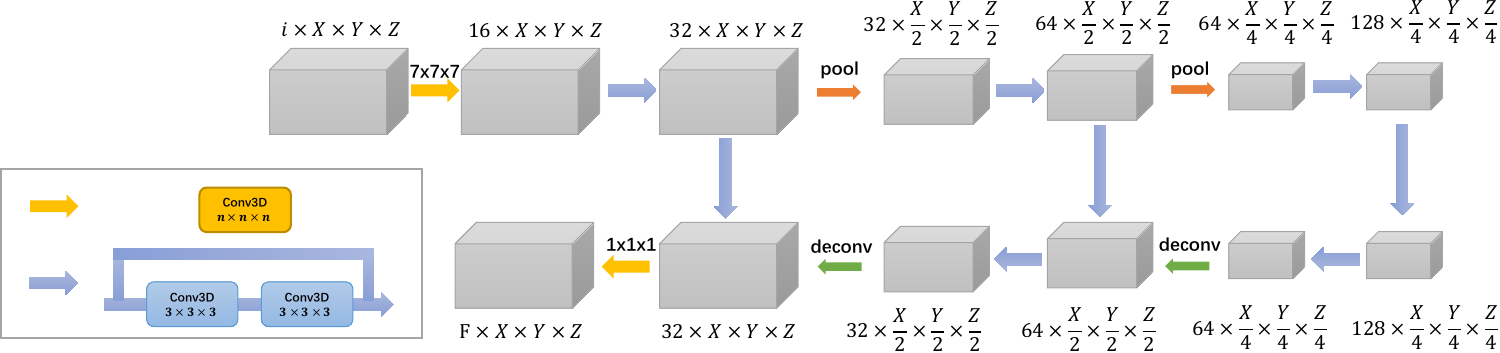}
   \caption{The structure and detailed setting of V2V-Net and Fusion-Net.}
   \label{fig:v2vnet}
\end{figure*}

Figure \ref{fig:v2vnet} shows the detailed structure of V2V-Net and Fusion-Net. $i=1$ for the one of point cloud information and $i=K$ for the one of RGB information, $K$ is the number of joints. $X, Y, Z$ represents the setting of volumetric space, and $F$ represents $F_1$ and $F2$. As indicated in the legend, the yellow arrow represents a standard 3D convolutional layer, while the blue arrow denotes a Residual Block consisting of two 3D convolutional layers. As indicated in the legend, the yellow arrow represents a standard 3D convolutional layer, while the blue arrow denotes a Residual Block consisting of two 3D convolutional layers.

% \begin{figure*}[!t]
%    \centering  
%    \includegraphics[width=1\linewidth]{figs/dataset2.png}
%    \caption{Real datasets and the corresponding synthetic datasets generated by SyncHuman.}
%    \label{fig:dataset}
% \end{figure*}

\section{Human Prior Loss}
   We designed the human prior loss to encourage the network to generate human-like 3D keypoints. The human prior loss comprises three components: 1) the predicted bone lengths should be within a reasonable range; 2) the predicted lengths of symmetric bones should be similar; and 3) the predicted bone angles should be reasonable according to human kinematics.

   We set a limited length range for all bones. In our case, 
   we set $l_{\text{min}}=0.05 \text{m}$ and $l_{\text{max}}=0.7 \text{m}$. So the $\mathcal{L}_{\text{length}}$ can be defined as:
   \begin{equation}
   \begin{aligned}
      \label{eq:length}
      \mathcal{L}_{\text{length}} = \sum_{b=1}^{N} \mathcal{C}(B_i-l_{\text{max}}, 0) + \mathcal{C}(l_{\text{min}}-B_i, 0), 
   \end{aligned}
   \end{equation}
   where $\mathcal{C}(\cdot)$ is the clipping function that clip the value greater than $0$, $N$ is the number of bones. 
   As to the symmetric bones, we set the symmetric bones as a pair, and set $L_{\text{2}}$ loss among them. So the $\mathcal{L}_{\text{symm}}$ can be defined as:
   \begin{equation}
   \begin{aligned}
      \label{eq:symm}
      \mathcal{L}_{\text{symm}} = \sum_{b=1}^{N} \Vert B_i - B_{\text{symm}(i)} \Vert_2,
   \end{aligned}
   \end{equation}

   \begin{figure}
      \centering
      \includegraphics[width=0.3\textwidth]{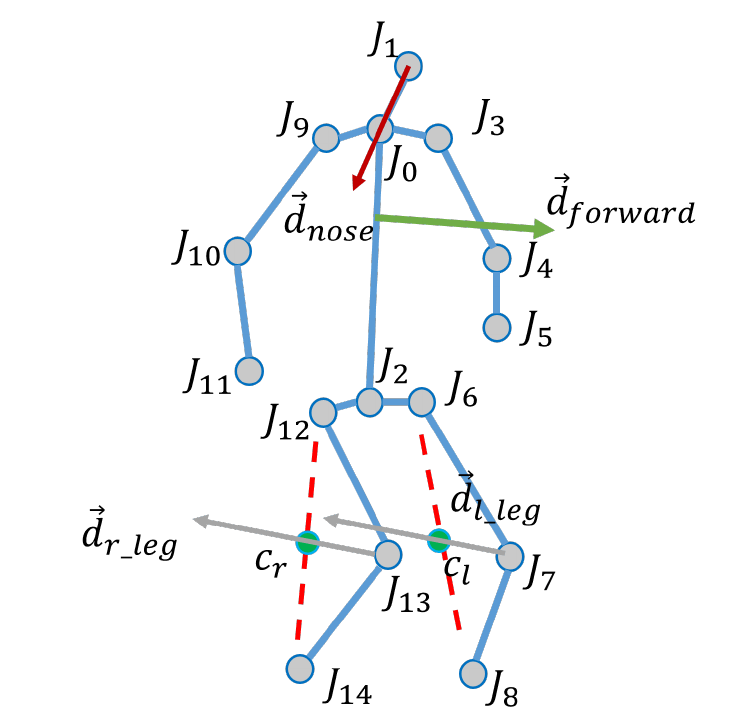}
      \caption{Definition of $\mathcal{L}_{\text{angle}}$.}
      \label{fig:pose angle}   
   \end{figure}     
   \noindent where $B_{\text{symm}}$ is the symmetric bone of $B_i$.
   As to angle loss, we limit the nose-neck-midhip angle and hip-knee-ankle angle specifically to let nose be in front of the 
   body and legs be bent forward. Figure \ref{fig:pose angle} shows the definition of each joint and vectors.
   Specially, we do not directly calculate the angle of the bones, but calculate the dot product of corresponding vectors.  
   First, we calculate the forward direction vector $\vec{d}_{\text{forward}}$ of the body, which is the cross product of the unit vector from neck to midhip $\overrightarrow{J_{\text{0}}J_{\text{2}}}$ and unit vector from 
   neck to left shoulder $\overrightarrow{J_{\text{0}}J_{\text{3}}}$:
   \begin{equation}
   \centering
   \label{eq:cross}
   \vec{d}_{\text{forward}} = \overrightarrow{J_{\text{0}}J_{\text{2}}} \times \overrightarrow{J_{\text{0}}J_{\text{3}}},
   \end{equation}
   Then, as to the nose-neck-midhip angle, 
   we calculate the unit vector from neck to nose $\overrightarrow{J_{\text{1}}J_{\text{0}}}$ denoted by $\vec{d}_{\text{nose}}$, 
   and we calculate the dot product of  $\vec{d}_{\text{nose}}$ and $\vec{d}_{\text{forward}}$ and get the head angle loss:
   \begin{equation}
   \begin{aligned}
      \label{eq:head}
      \mathcal{L}_{\text{head\_ang}} = \mathcal{C}(\vec{d}_{\text{forward}} \cdot \vec{d}_{\text{nose}}, 0, 1),
   \end{aligned}
   \end{equation}
   where $\mathcal{C}(\cdot)$ is the clipping function that clip the value into $0$ to $1$.
   As to the hip-knee-ankle angle, we need to get the midpoint of the hip and ankle denoted by $c_{\text{l}}$ and 
   $c_{\text{r}}$ for left leg and right leg respectively. Then, we calculate the unit vectors from knee point to the leg's midpoint 
   as $\vec{d}_{\text{l\_leg}}$ and $\vec{d}_{\text{r\_leg}}$. Therefore, we get the leg angle loss:
   \begin{equation}
   \begin{aligned}
      \label{eq:leg}
      \mathcal{L}_{\text{leg\_ang}} = \mathcal{C}(\vec{d}_{\text{forward}} \cdot \vec{d}_{\text{l\_leg}}, 0, 1) + 
            \mathcal{C}(\vec{d}_{\text{forward}} \cdot \vec{d}_{\text{r\_leg}}, 0, 1),
   \end{aligned}
   \end{equation}
   where $\mathcal{C}(\cdot)$ is the clipping function that clip the value into $0$ to $1$.
   Therefore, we can calculate the angle loss:
   \begin{equation}
   \begin{aligned}
      \label{eq:angle}
      \mathcal{L}_{\text{angle}} = \mathcal{L}_{\text{head\_ang}} + \mathcal{L}_{\text{leg\_ang}},
   \end{aligned}
   \end{equation}
   Finally, we combine the three losses together as the human prior loss:
   \begin{equation}
   \begin{aligned}
      \label{eq:bone}
      \mathcal{L}_{\text{prior}} = \gamma_{\text{1}}\mathcal{L}_{\text{length}} + \gamma_{\text{2}}\mathcal{L}_{\text{symm}} + \gamma_{\text{3}}\mathcal{L}_{\text{angle}},
   \end{aligned}
   \end{equation}
   where $\gamma_{\text{1}}$, $\gamma_{\text{2}}$ and $\gamma_{\text{3}}$ are the weights of each loss. In our case, we set 
   all the weights as $1$.

\begin{table}[!t]
   \centering
   \caption{Human detection results on different datasets. ``$\#$'' means using synthetic datasets (BaseketBallSync and PanopticSync) to pretrain and 
      directly evaluate on corresponding real-world datasets.}
   \smallskip
   \label{tab:det}
   \begin{tabular}{c c cc}
   \toprule
   \multirow{2}{*}{\textbf{Datasets}}   & \multirow{2}{*}{\textbf{Methods}} & \multicolumn{2}{c}{\textbf{Metric}} \\ \cmidrule(lr){3-4}
                                 &                          & $\text{AP}_{\text{50}}$        & $\text{AP}_{\text{70}}$        \\ \midrule
   \multirow{3}{*}{BasketBall} & MVDet                    &  69.41       &  37.66           \\
                                 & PointPillars$\#$            & 88.17        & 44.26            \\
                                 & PointPillars             & \textbf{89.77}        & \textbf{69.96}            \\ \midrule
   \multirow{3}{*}{Panoptic}   & VoxelPose            & 21.17             & 0.19            \\
                                 & PointPillars$\#$            & 40.17             & 6.25            \\
                                 & PointPillars                & \textbf{73.83}             & \textbf{13.97}            \\ \bottomrule
   \end{tabular}
\end{table}

\section{Extended Experiments}
   In this section, we conduct experiments to verify the advantages of using point cloud input for pedestrian detection. Additionally, we present more examples to explain the relationship between entropy value and pose rationality.

\begin{figure}[!t]
   \includegraphics[width=0.4\textwidth]{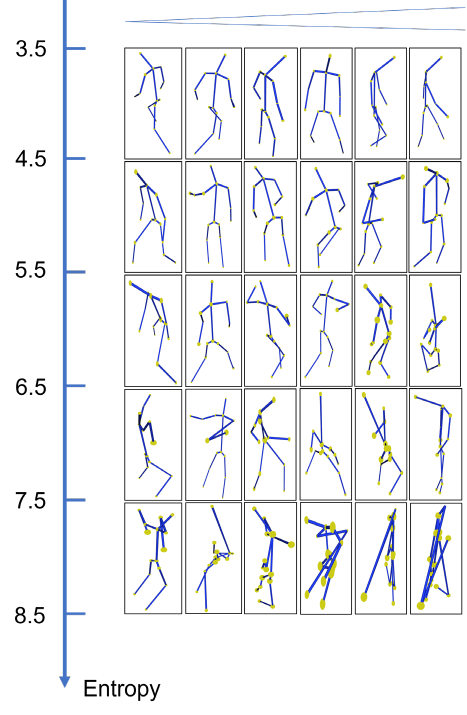}
   \caption{The entropy value and the specific poses. \includegraphics[height=0.2cm]{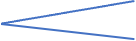} represents an increasing entropy value from left to right among row's samples. 
   The size of joints' ball represents the magnitude of the joint's entropy value.}
   \label{fig:entropy2}
\end{figure}

\subsection{Human Detection}
   For evaluating human detection, we assess performance using the established average precision (AP) metric as described in KITTI \cite{geiger2012we}. We consider detections as true positives if they overlap by more than 70\% (AP\(_{70}\)) or 50\% (AP\(_{50}\)).

   % In the application of our current experiment, we adopt PointPillars \cite{pointpillars} to detect the human bounding box. And we utilize the VoxelPose's CPN \cite{2020voxelpose} and MVDet \cite{mvdet} which is more suitable 
   % in large scene's application as for comparison in terms of multi-view RGB-based methods. In CMU panoptic studio setup, VoxelPose \cite{2020voxelpose} can get relatively accurate center location. However, it set the size of bounding box to be 
   % a constant value (We set 0.8m $\times$ 0.8m $\times$ 1.9m in our case for tighting results instead of 2m $\times$ 2m $\times$ 2m in \cite{2020voxelpose}) which affects the performance of detection. In BasketBall, we adopt MVDet to detect human beings. Table \ref{tab:det} shows that 
   % pointcloud-based method outperforms the multi-view RGB-based method in terms of $\text{AP}_{\text{50}}$ and $\text{AP}_{\text{70}}$ benefiting from original poincloud's 3D information. Besides, we verify the generalization ability of the pointcloud-based method by pretraining it on 
   % our synthetic dataset, and it still has an acceptable result.
   In our current experiment, we adopt PointPillars \cite{pointpillars} to detect human bounding boxes. For comparison with multi-view RGB-based methods, we utilize VoxelPose's CPN \cite{2020voxelpose} and MVDet \cite{mvdet}, which is more suitable for large scene applications. In the CMU Panoptic Studio setup, VoxelPose \cite{2020voxelpose} achieves relatively accurate center localization. However, it sets the bounding box size to a constant value (we use 0.8m \(\times\) 0.8m \(\times\) 1.9m for tighter results, compared to 2m \(\times\) 2m \(\times\) 2m in \cite{2020voxelpose}), which affects detection performance. In the BasketBall dataset, we adopt MVDet to detect humans. Table \ref{tab:det} shows that the point cloud-based method outperforms the multi-view RGB-based method in terms of AP\(_{50}\) and AP\(_{70}\), benefiting from the 3D information of the original point cloud. Additionally, we verify the generalization ability of the point cloud-based method by pretraining it on our synthetic dataset, and it still produces acceptable results.

\begin{figure*}[!t]
   \centering
   \includegraphics[width=0.9\linewidth]{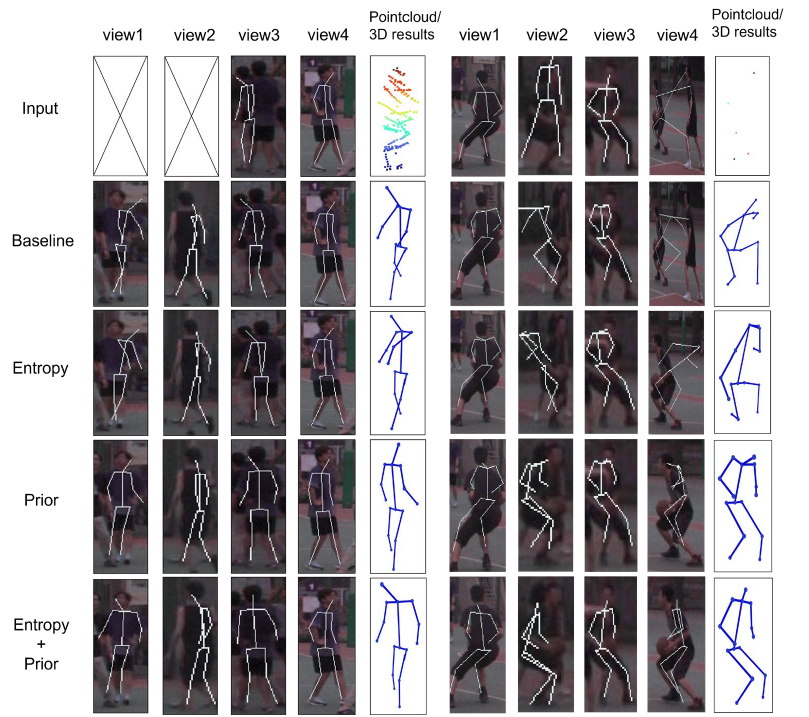} 
   \vspace{-0.1cm}
   \caption{Qualitative visualization on BasketBall about different unsupervised training losses. ``Baseline'' uses only pseudo 2D pose supervision. ``Entropy'' indicates the addition of entropy-selected pseudo 3D pose supervision. ``Prior'' denotes the incorporation of human prior loss.}
   \label{fig:basketball2}
\end{figure*}

\subsection{The analysis of unsupervised training losses.}
   Figure \ref{fig:basketball2} qualitatively shows that these designed unsupervised training losses significantly enhance robustness to 2D pose estimation errors.

\subsection{Entropy Analysis}
   Figure \ref{fig:entropy2} shows the entropy value and the specific poses, and we can find that the 3D poses become more and more irrational while the entropy goes up.

\end{document}